\documentclass[pdflatex,sn-basic]{sn-jnl}% Basic Springer Nature Reference Style/Chemistry Reference Style

\usepackage{cleveref}
\usepackage{subfigure}
\usepackage{tikz}
\usepackage{program}
\usepackage{bm}

\crefname{section}{§}{§§}
\Crefname{section}{§}{§§}

\usepackage{etoolbox}

\jyear{2021}%

\theoremstyle{thmstyleone}%
\theoremstyle{thmstyletwo}%

\theoremstyle{thmstylethree}%

\begin{document}

\title[]{Visual Knowledge in the Big Model Era: Retrospect and Prospect}
%\title[Article Title]{Machine learning for retrosynthesis: generating reactions from millions of compounds}

\author[1]{\fnm{Wenguan} \sur{WANG}}\email{wenguanwang@zju.edu.cn}

\author[1]{\fnm{Yi} \sur{YANG}}\email{yiyang@zju.edu.cn}

\author[1]{\fnm{Yunhe} \sur{PAN}}\email{panyh@zju.edu.cn}
% %\equalcont{These authors contributed equally to this work.}
% %\equalcont{These authors contributed equally to this work.}

%\author[1]{\fnm{First} \sur{Last}}\email{examples@zju.edu.cn}

\affil[1]{\orgdiv{College of Computer Science and Technology}, \orgname{Zhejiang University}, \orgaddress{\city{Hangzhou}, \postcode{310027}, \state{Zhejiang}, \country{China}}}


\abstract{
    Visual knowledge is a new form of knowledge representation that can encapsulate visual concepts and their relations in a succinct, comprehensive, and interpretable manner, with a deep root in cognitive psychology. As the knowledge about the visual world has been identified as an indispensable component of human cognition and intelligence, visual knowledge is poised to have a pivotal role in establishing machine intelligence. With the recent advance of Artificial Intelligence (AI) techniques, large
    AI models (or foundation models) have emerged as a potent tool capable of extracting versatile patterns
    from broad data as implicit knowledge, and abstracting them into an outrageous amount of numeric parameters. To pave the way for creating visual knowledge empowered AI machines in this coming wave, we present a timely review that investigates the origins and development of visual knowledge in the pre-big model era, and accentuates the opportunities and unique role of visual knowledge in the big model era.}




\maketitle

\section{Introduction} \label{sec:introduction}

$_{\!}$The$_{\!}$ concept$_{\!}$ of$_{\!}$ visual$_{\!}$ knowledge$_{\!}$~\citep{pan2019visual}$_{\!}$ was recently proposed as a form of knowledge representation that differs from the traditional ones adopted/learned by symbolic and sub-symbolic AI approaches (\textit{e.g.}, knowledge graph, handcrafted image descriptors, distributed$_{\!}$ visual$_{\!}$ representations).$_{\!}$ Drawing$_{\!}$ on cognitive$_{\!}$ studies$_{\!}$~\citep{anderson1980cognitive}$_{\!}$ of human mental imagery, which enables us to manipulate$_{\!}$ visual$_{\!}$ entities$_{\!}$ in$_{\!}$ our$_{\!}$ mind,$_{\!}$ visual$_{\!}$ knowledge$_{\!}$ theory$_{\!}$ posits$_{\!}$ that$_{\!}$ next-generation$_{\!}$ AI$_{\!}$ needs$_{\!}$ to$_{\!}$ fully$_{\!}$ express$_{\!}$ visual$_{\!}$ concepts$_{\!}$ and$_{\!}$ their$_{\!}$ attributes$_{\!}$ (\textit{e.g.},$_{\!}$ shape, structure, motion, affordance), as well as reason about their transformations, compositions, comparisons, predictions, and narrations, through a unified, abstract, and interpretable form of representation.

$_{\!}$After$_{\!}$ the$_{\!}$ emergence$_{\!}$ of$_{\!}$ large$_{\!}$ language$_{\!}$ models$_{\!}$~like$_{\!}$
GPT-3~\citep{brown2020language}, the field of natural language$_{\!}$ processing$_{\!}$ has$_{\!}$ experienced$_{\!}$ remarkable$_{\!}$ advancements: traditional ``narrow'' language models that are trained to perform specific tasks in a single domain are giving way to highly sophisticated and versatile language models that are trained on a vast corpus of unlabeled textual data that can be used for different language tasks across domains. Like GPT for natural language processing, the recent work known as Segment Anything Model (SAM)~\citep{kirillov2023segment} ushered the field of computer vision into the era$_{\!}$ of$_{\!}$ visual$_{\!}$ foundation$_{\!}$ models$_{\!}$ ---$_{\!}$ by$_{\!}$ training$_{\!}$ on$_{\!}$ $>$1B segmentation masks in $>$11M natural images, SAM shows the promise of a broad applicability to various image$_{\!}$ segmentation$_{\!}$ tasks,$_{\!}$ without$_{\!}$ re-training$_{\!}$~or fine-tuning as previously needed. With incredible speed, large models are revolutionizing AI field and transforming the landscape of scientific research.

Albeit the unprecedented progress, it is becoming increasingly evident that large
AI models still suffer several deficiencies that compromise their reliability and efficacy. Chief among these are their pronounced opacity, which poses great challenges for trust, accountability, and effective debugging, as well as their insatiable demand for data and computational resources, which raise both ethical and environmental concerns. These limitations are inherited from their rudimentary predecessors but exacerbated by their heightened sophistication and scale. Compounding these concerns,   large AI models are susceptible to generating nonsensical or unfaithful content, known as ``hallucination'', exposing their inherent biases, lack of real-world understanding, and weakness in generalizing or reasoning beyond their scope.

\begin{figure*}[t]
	\centering
	\includegraphics[width=\linewidth]{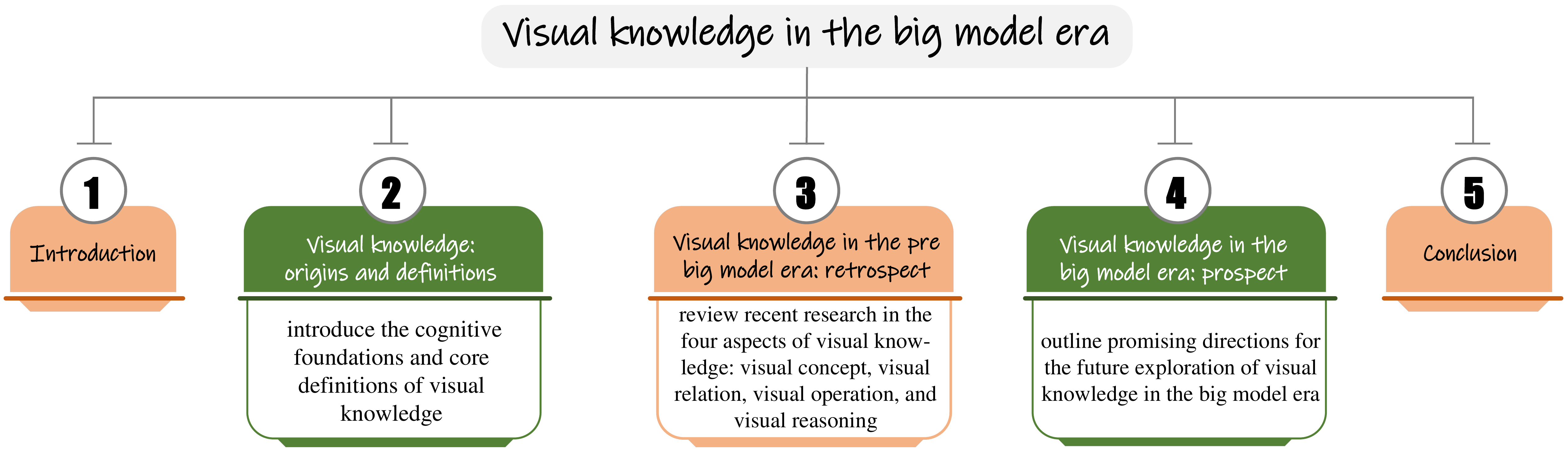}
	\caption{The overall structure of this article. } \label{fig:overview}
\end{figure*}

Considering the appealing advantages of visual knowledge in terms of expressive and interpretable representation, manipulation, and reasoning of visual concepts, it is probably fair to assume that the weaknesses of big AI models can be (at least in part) alleviated through a deeper understanding and development of visual knowledge. On the other hand, given the tremendous success of big AI models and the significant challenges of visual knowledge acquisition, it appears to be apparent that future endeavors should be made to research specific techniques which seek to building visual knowledge with the aid of large-scale statistical$_{\!}$ learning.$_{\!}$ These$_{\!}$ are$_{\!}$ the$_{\!}$ considerations$_{\!}$~that$_{\!}$ give rise to the work presented here. We delve into early theoretical and methodological investigations of visual knowledge, and demonstrate that the key insights provided by visual knowledge studies are shown to shed new light onto the increasingly prominent role of trust, interpretability, and accountability in the new AI revolution sparked by big models. Moreover, we identify promising directions for the creation of more powerful AI systems that harness the synergies between visual knowledge and big models to overcome the weaknesses of each other. This also turns out that the development of new, deeper forms of visual knowledge in large-scale charts the course for next-generation AI~\citep{pan2020multiple,yang2021multiple}.

%In this work, we position previous research around visual knowledge in the context of the recent explosion of interest and excitement about big AI models.

The following sections are organized as follows (Fig.~\ref{fig:overview}): we first briefly introduce the origin and theory of visual$_{\!}$ knowledge$_{\!}$ (Sec.\!~\ref{sec:ob}).$_{\!}$ Subsequently,$_{\!}$ we$_{\!}$ systematically categorize latest advances in visual knowledge based on a taxonomy from four different perspectives including visual concept, visual relation, visual operation, and visual reasoning, which are key elements and characteristics of visual knowledge (Sec.\!~\ref{sec:re}). Based on the analysis of the research situation of visual knowledge, we outline a series of promising directions that can act as a compass for future explorations of visual knowledge in the imminent age of big models (Sec.\!~\ref{sec:pr}). We conclude this work in Sec.\!~\ref{sec:con}. We hope our efforts in this paper can bring together computer vision, graphics, and machine learning communities as well as industry to advance further development of visual knowledge, and inch the level of general intelligence in machines closer to that of humans.

\section{Visual knowledge: origins and definitions}\label{sec:ob}

To facilitate a comprehensive understanding of visual knowledge theory, in this section, we delve into its cognitive foundations (Sec.~\ref{sec:21}) and core definitions (Sec.~\ref{sec:22}).

%Memory is not a single unitary system, but rather a complex set of processes that involve different brain regions and functions. One way to classify memory is based on the duration of storage: sensory memory, short-term memory, and long-term memory. Another way to classify memory is based on the type of information stored: declarative memory (facts and events) and non-declarative memory (skills and habits).

\subsection{Origins}\label{sec:21}
% Gilster’s idea of digital literacy did not appear “out of the blue.” h ere was already a substantial set of literature and practical experience around the ideas of information literacy and computer literacy:
The theory of visual knowledge~\citep{pan2019visual} did not appear ``out of the blue.'' Rather, its roots extend deeply into the realm of cognitive psychology. \\

\noindent\textbf{The significant role of visual signals in information processing of human brain}

Our knowledge of the world is not solely derived from textual and verbal material, but also from the visual perception of the real environment. Pioneering research in cognitive and biological psychology~\citep{milner2006visual} have revealed that nearly half of our cerebral cortex is dedicated to processing visual stimuli. In addition, the human brain processes images $60,000$ times faster than text, and $90$ percent of information transmitted to the brain is visual. These statistics suggest that the human brain places a high value on visuals over any other type of information.\\

%Just as our work investigates relations between visual media and non-visual formations, it also concentrates on relations across different visual media and on the ways that visual objects become irreducible to language or text.
%our memory can retain much larger sets of pictures, while only retain a few auditory items.

\noindent\textbf{Visual memory:$_{\!}$ capacity, function, and representation of the stored content}

Recent cognitive research also provided strong evidence that human memory for pictures is much better than memory for sounds~\citep{bigelow2014achilles}. Human \textit{visual memory}$_{\!}$ is$_{\!}$ ubiquitous$_{\!}$ in$_{\!}$ daily$_{\!}$ life$_{\!}$ and$_{\!}$ closely$_{\!}$ links$_{\!}$ to$_{\!}$~many$_{\!}$ high-level$_{\!}$ cognitive$_{\!}$ functions,$_{\!}$ such$_{\!}$ as$_{\!}$ \textit{mental imagery} (the$_{\!}$ ability$_{\!}$ to$_{\!}$ create$_{\!}$ an$_{\!}$ image$_{\!}$ in$_{\!}$ mind$_{\!}$ in$_{\!}$ the absence$_{\!}$ of$_{\!}$ sensory$_{\!}$ input).$_{\!}$ Visual$_{\!}$ memory,$_{\!}$ actively (\textit{i.e.}, visual working memory) or passively (\textit{i.e.}, visual long-term memory), holds and recalls visual information in mind, making it accessible and manipulable in support of ongoing cognitive tasks. Rather than solely concentrating on the capacity and cognitive function of visual memory, cognitive psychologists endeavored to explore the representation of the content stored in visual memory. \cite{shepard1988mental,shepard1972chronometric,kosslyn1978visual,moyer1973comparing} conducted a series of experiments showing that visual memory representations,  in contrast to verbal memory representations, support a variety~of mental manipulations, including rotation, folding, scanning, and analogizing. Cognitive psychologists  also found evidence suggesting that the structure of visual memory representations can be thought of as hierarchically organized~\citep{brady2011review}.\\

\noindent\textbf{Interactions among perception, visual memory, and human knowledge}

Compared with verbal memory that is primarily processed in the left hemisphere, visual memory tends to be more bilateral. Visual memory can be \textit{episodic} (\textit{i.e.}, memory of visual events or experiences that have a specific time and place), but it can also be \textit{semantic} (\textit{i.e.}, memory of general facts or visual concepts that are not tied to a specific context). Human store knowledge about most items in the real world and there is clear evidence that the representation of content in visual memory is not just a straightforward recording of sensory input but depends upon our past experience and our stored knowledge~\citep{brady2011review}. %Human reasoning has consistent patterns across various tasks~\citep{sternberg1986toward}. This unity becomes apparent through cross-task studies. Many theorists believe shared processes underlie these patterns, especially in inductive and deductive reasoning tasks~\textit{e.g.}, \citep{greeno1978natures, pellegrino1979cognitive}.
Here it is necessary to make the distinction between visual memory and human stored knowledge. Visual memory refers to the ability to remember visual information that was seen previously. Thus, visual memory is the storage and subsequent retrieval of  perceived visual information. Stored knowledge refers to the preexisting representations that underlie our ability to recognize and understand visual input. For example, when we first view an image, say of an orange, stored knowledge about the visual form and features of oranges in general enables us to recognize the object as such. Later, if we encounter another picture of an orange, visual memory enables us to decide whether it is the exact same orange we saw previously. Thus, specific items for which we have expertise, like faces, are represented with more fidelity~\citep{scolari2008perceptual}, while general concepts are represented after statistical regularities~\citep{brady2008compression}. Cognitive research  has also shown that our stored knowledge can modulate how we form and utilize mental images and visual memory, and our knowledge and visual memory can affect how we perceive and attend to visual stimuli.\\

%By ``visual memory,'' we refer to the ability to explicitly remember an image that was seen previously but that has not been continuously held actively in mind. Thus, visual memory is the storage and subsequent retrieval of  visually perceived information. By ``stored knowledge,'' we refer to the preexisting visual representations that underlie our ability to perceive and recognize visual input. For example, when we first see an image, say of a red apple, stored knowledge about the visual form and features of apples in general enables us to recognize the object as such. If we are shown another picture of an apple later, visual memory enables us to decide whether this is the exact same apple we saw previously.

%Before discussing the capacity of long-term memory, it is important to make the distinction between visual long-term memory and stored knowledge.

%Thus, items for which we have expertise, like faces, are represented with more fidelity (Curby  Gauthier, 2007; Scolari et al., 2008). and more individual colors can be represented after statistical regularities between those colors are learned In addition, the representation of individual items are biased by past experience (e.g., Huang  Sekuler, 2010; Huttenlocher et al., 2000). the representation of even simple items in working memory depends upon our past experience with those items and our stored visual knowledge.

\noindent\textbf{The proposal of visual knowledge theory}

The aforementioned cognitive studies evidenced the close and complex relations between visual information processing, visual memory, and human knowledge: the information gleaned from visual experiences supports many cognitive functions (such as visual memory and mental imagery) as well as the construction of our knowledge; and such knowledge, in turn, shapes visual memory, influences perception, and greatly facilitates our understanding of the world around us. %Through the reasoning process, it also boosts an intelligent agent's comprehension and execution capabilities for tasks.
Taking all these into consideration, a plausible argument can be made that one of the shortcomings of existing AI research is the scarcity of studies concerning human mental representations about visual items. The theory of visual knowledge~\citep{pan2019visual} is thus arising to fill such gap. %under such a background.

\subsection{Definitions}\label{sec:22}
Basically, our visual knowledge are stable mental representations about visual objects and the commonalities in the inherent rules among various tasks. They are abstracted from our visual experiences and memory, and retained in our mind. They enable ours to remember, imagine, and reason about the world and accomplish targeted tasks. Neuropsychological investigations have also revealed some characteristics of our mental representations of visual objects:
\begin{itemize}%[leftmargin=*]
   \setlength{\itemsep}{0pt}
   \setlength{\parsep}{-2pt}
   \setlength{\parskip}{-0pt}
   \vspace{-4pt}
   \item The$_{\!}$ ability$_{\!}$ to$_{\!}$ capture$_{\!}$ the$_{\!}$ typical$_{\!}$ attributes of visual objects, such as their shapes, sizes, colors, and textures.
   \item The ability to describe the static and dynamic relationships between visual objects, such as  relative positions,  actions, velocities, and temporal sequences.
   \item The ability to perform spatio-temporal operations over visual objects, such as transforming shapes, actions, and scenes, making analogies and associations, and predicting future outcomes.
    \item The ability to engage in reasoning, such as analogizing, inducting, and deducing new tasks, combining existing concepts to form new concepts, and generalizing from anomalous samples.
   \vspace{-4pt}
\end{itemize}

Visual knowledge is not just an abstract representation of visual objects, but involves an active and generative process that supports various cognitive skills. Hence one of the core insights of visual knowledge theory is that AI systems should also develop and use visual knowledge in a similar way.

More specifically, visual knowledge~\citep{pan2019visual}, as a new form of knowledge representation, is constructed as a combination of four essential components, namely visual concept (Sec.~\ref{sec:221}), visual relation (Sec.~\ref{sec:222}), visual operation (Sec.~\ref{sec:223}), and visual reasoning (Sec.~\ref{sec:224}). With these key components, visual knowledge can enable AI systems to comprehensively describe, robust recognize, and reason about visual items and solve tasks.

\subsubsection{Visual concept}\label{sec:221}

\begin{figure}[tbh]
	\centering
	\includegraphics[width=0.6\linewidth]{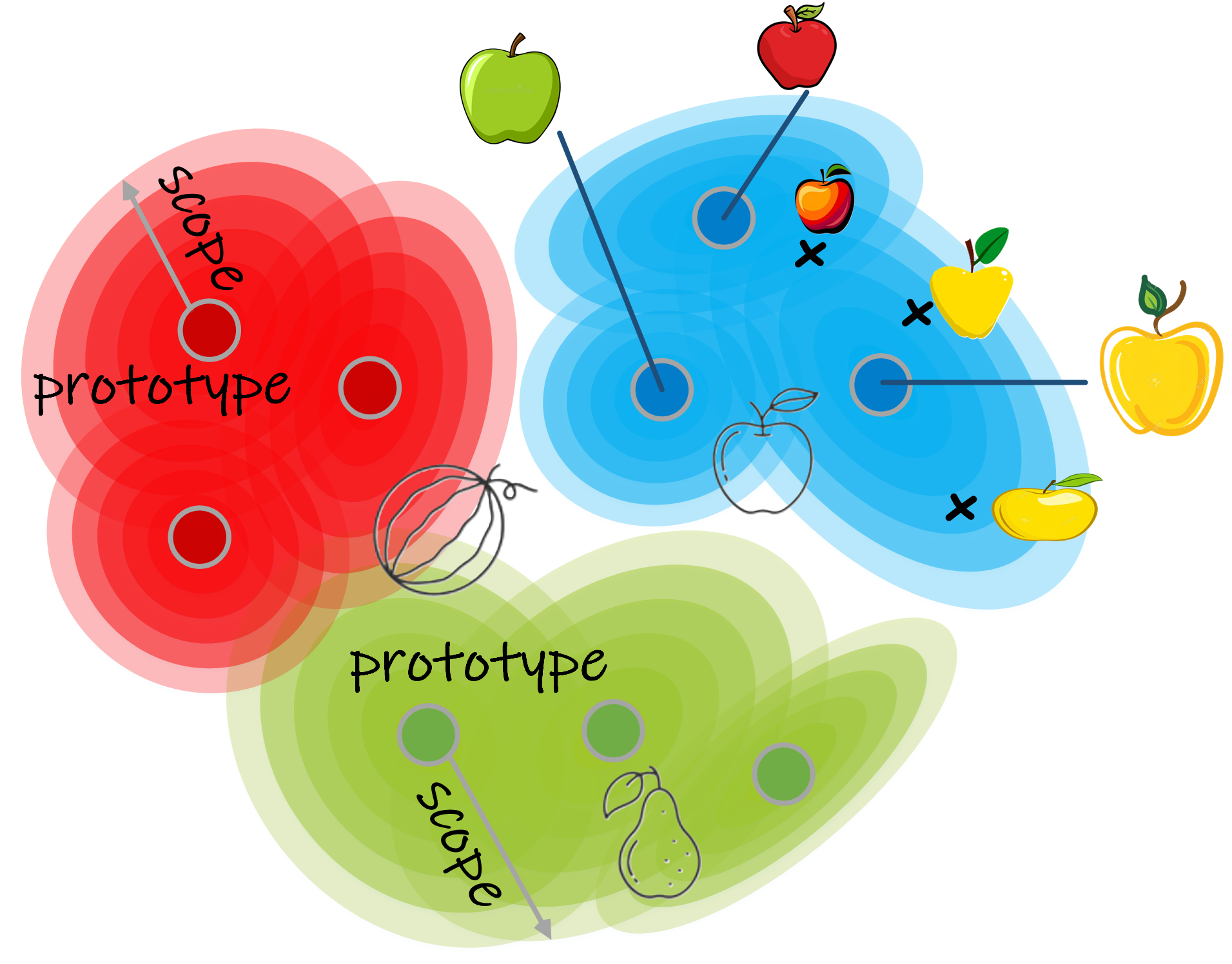}
	\caption{Illustration of prototype-and scope-based visual concept representation. Here we show three visual concepts, namely \textit{pear}, \textit{apple}, and \textit{watermelon}. } \label{fig:visual_concept}
\end{figure}

A visual concept is a category of visual objects that share some common features. Visual knowledge theory holds that a visual concept is defined by \textit{prototype} and$_{\!}$ \textit{scope}.$_{\!}$ When$_{\!}$ thinking$_{\!}$ of$_{\!}$ a$_{\!}$ visual$_{\!}$ concept, such as apple, we form a mental set
of images that represent the most common or typical features/attributes of that concept.$_{\!}$ These$_{\!}$ images$_{\!}$ are$_{\!}$ called$_{\!}$ prototypes,$_{\!}$ serving as the basis for generating or recognizing any variation of that concept. For instance,  we might have prototypes of apples that are red, green, or yellow, and that are round, oval, or heart-shaped. Based on these prototypes, we can imagine or identify any apple that has similar features, even if it is not exactly the same as$_{\!}$ any$_{\!}$ of$_{\!}$ the$_{\!}$ prototypes.$_{\!}$ But$_{\!}$ apples$_{\!}$~do$_{\!}$~not look$_{\!}$ exactly$_{\!}$ the$_{\!}$ same;$_{\!}$ some$_{\!}$ might$_{\!}$ be$_{\!}$ lighter$_{\!}$ or$_{\!}$ darker, bigger or smaller, smoother or rougher than others. However, there exists a limit or boundary to how much an apple can deviate from the prototypes and still be considered an apple. If the shape or color is too different, it might belong to another visual concept, such as \textit{pear} or \textit{watermelon}. The range of variation that is acceptable for a category is called scope. Shapes and colors within the scope are considered part of the category of apples, but shapes and colors outside it are not (see Fig.~\ref{fig:visual_concept}).

The idea of using prototype to represent visual concepts is in line with the classic \textit{prototype theory}, which in large part owes its beginnings to~\cite{rosch1975family}, and has gained widespread recognition in cognitive science and other fields. Prototype theory provides an important theoretical account of cognitive categorization. It posits that a category of things in the world (objects, animals, shapes, \textit{etc}.) can be represented in the mind by a prototype. A prototype is a cognitive representation that captures the regularities and commonalities among category members. According to
prototype theory, objects are classified by first comparing them to the prototypes stored in memory, evaluating the similarity evidence from those comparisons, and then
classifying the item in accord with the most similar prototype. Formally, let $\mathcal{X}$ be the data space and $\mathcal{Y}\!=\!\{y_1,_{\!} \cdots_{\!}, y_C\}_{\!}$ the set of $C$ categories. Given an data instance $\bm{x}\in\mathcal{X}$, a prototype classification model assigns
it to the class $y\in\mathcal{Y}$ with the closest prototype:
\begin{equation}
{y}\!=\!\mathop{\arg\min}\nolimits_{y_c\!\!~\in\!\!~\mathcal{Y}} \langle\bm{x}, {\bm{p}}_c\rangle,
\end{equation}
where $\langle\cdot,\cdot\rangle$ is a distance measure, ${\bm{p}}_c$ refers to the prototype of category $y_c$, and a certain dimension of $\bm{x}$ ($\bm{p}$) encodes a specific salient attribute.  This novel prototype view had enabled researchers to develop many computational models for categorization, including the famous $k$-Nearest$_{\!}$ Neighbors$_{\!}$ ($k$-NN)$_{\!}$ and Nearest Centroids~\citep{fix1952discriminatory,cover1967nearest}. These prototype models differ mainly in how the prototypes are derived. For instance, the $k$-NN algorithm treats the $k$-nearest neighbors of the data samples as prototypes, while the Nearest Centroids algorithm considers the centroid, or average, of each category as the prototype.

In general, prototype theory is well-suited to explain the learning of many visual categories with a strong family-resemblance structure. However, \textbf{prototype theory lacks the notion of scope}, making it less tolerant to intra-class variance.

From a statistical perspective, the use of prototype and scope to describe category is essentially to capture the form or structure of the data distribution $p(x\!\mid\!y)$, \textit{i.e.}, how the data samples of a certain category look like. Hence the computational model of visual concept based categorization is a \textit{generative classifier}, which estimates the conditional probability of a label given an input, and then uses Bayes' rule to assign the most likely label~\citep{mackowiak2021generative}:
  \begin{equation}\small
  \begin{aligned}\label{eq:gc}
      p(y\!\mid\!\bm{x}) \!=\! \frac{p(y)p(\bm{x}\!\mid\!y)}{\sum_{c\!\!~\in\!\!~\mathcal{Y}}p(c)p(\bm{x}\!\mid\!c)}.
  \end{aligned}
\end{equation}
Unlike \textit{discriminative classifiers}, which directly map inputs to labels without explicit modeling of data distribution, generative classifiers are more difficult to train, because they have to model more aspects of the data~\citep{DBLP:conf/nips/LiangWM022}. This partially explains the challenges of constructing visual knowledge.

\subsubsection{Visual relation}\label{sec:222} %spatiotemporal relations (such as spatial location, orientation, size, distance, \textit{etc}.),
The term ``visual relation'' in visual knowledge theory denotes the connections and interactions that prevail among visual concepts, which are pivotal in navigating the complex landscape of visual cognition. Humans harbor an extensive repository of knowledge pertaining to the attributes of visual objects, a repository that transcends the mere intrinsic properties of these objects -- such as color, shape, and texture -- to include the relational properties interlinking them. These encompass their relative positioning, semantic dependencies, and affordances, which collectively constitute the relational properties, or visual relations. These relational properties, or visual relations,  are amenable to categorization into distinct classes, each illuminating different facets of visual cognition:

\begin{figure}[tbh]
	\centering
	\includegraphics[width=0.8\linewidth]{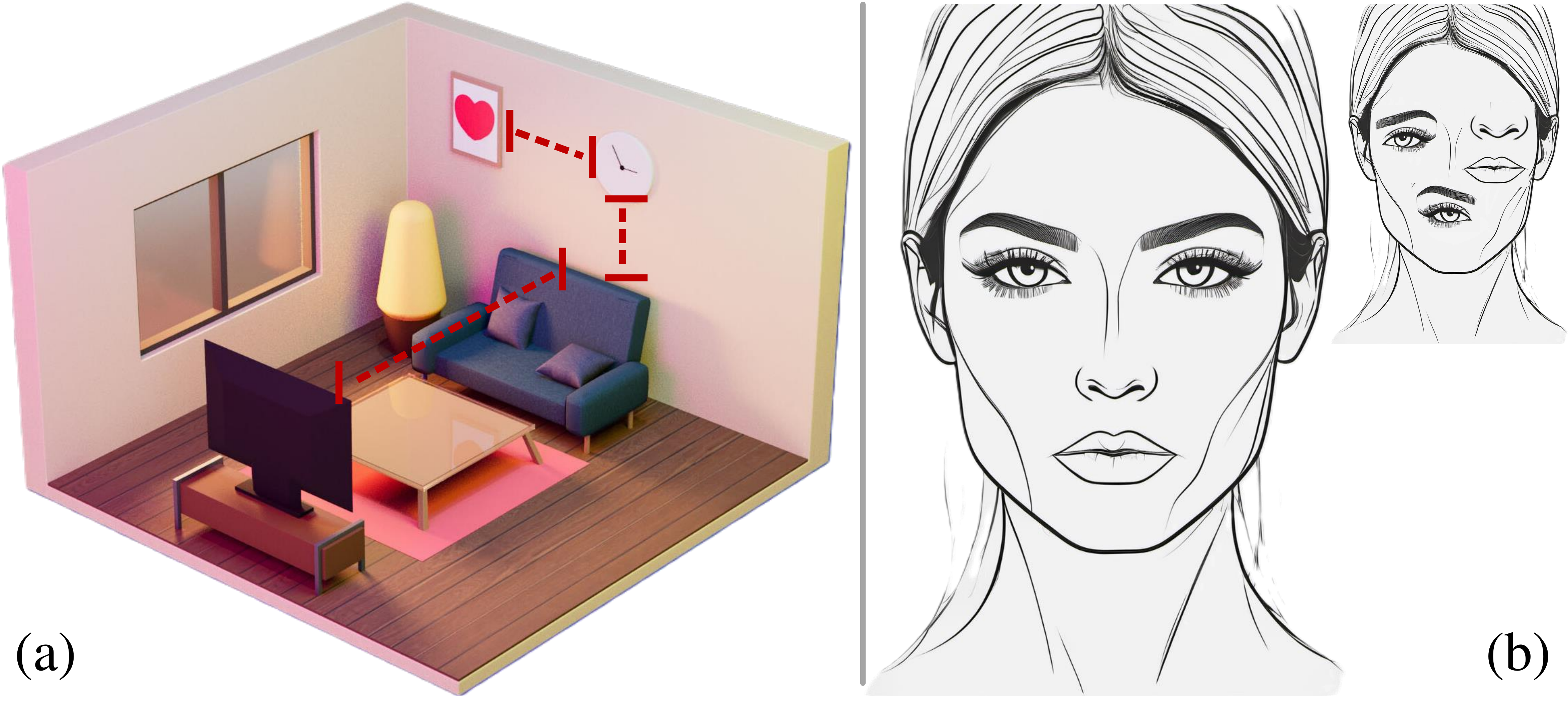}
	\caption{Illustration of geometric relations.} \label{fig:georelation}
\end{figure}

\noindent\textbf{Geometric relations:} These relations delineate how objects or concepts are interconnected based on their spatial configurations and geometric constructs, such as their relative position, direction, distance, intersection, alignment, parallelism, perpendicularity, and so on (see Fig.~\ref{fig:georelation}(a)). Such relations facilitate an understanding of the structure and organization of objects within the environment, and unveil the inherent harmony and order of nature and art. For example, the kernels in an apple are located at its center.  Similarly, our knowledge about the human face is not only the occurrence of eyes, nose, mouth, and ears, but also the precise spatial arrangement of those key facial elements (Fig.~\ref{fig:georelation}(b)).

\begin{figure}[tbh]
	\centering
	\includegraphics[width=0.8\linewidth]{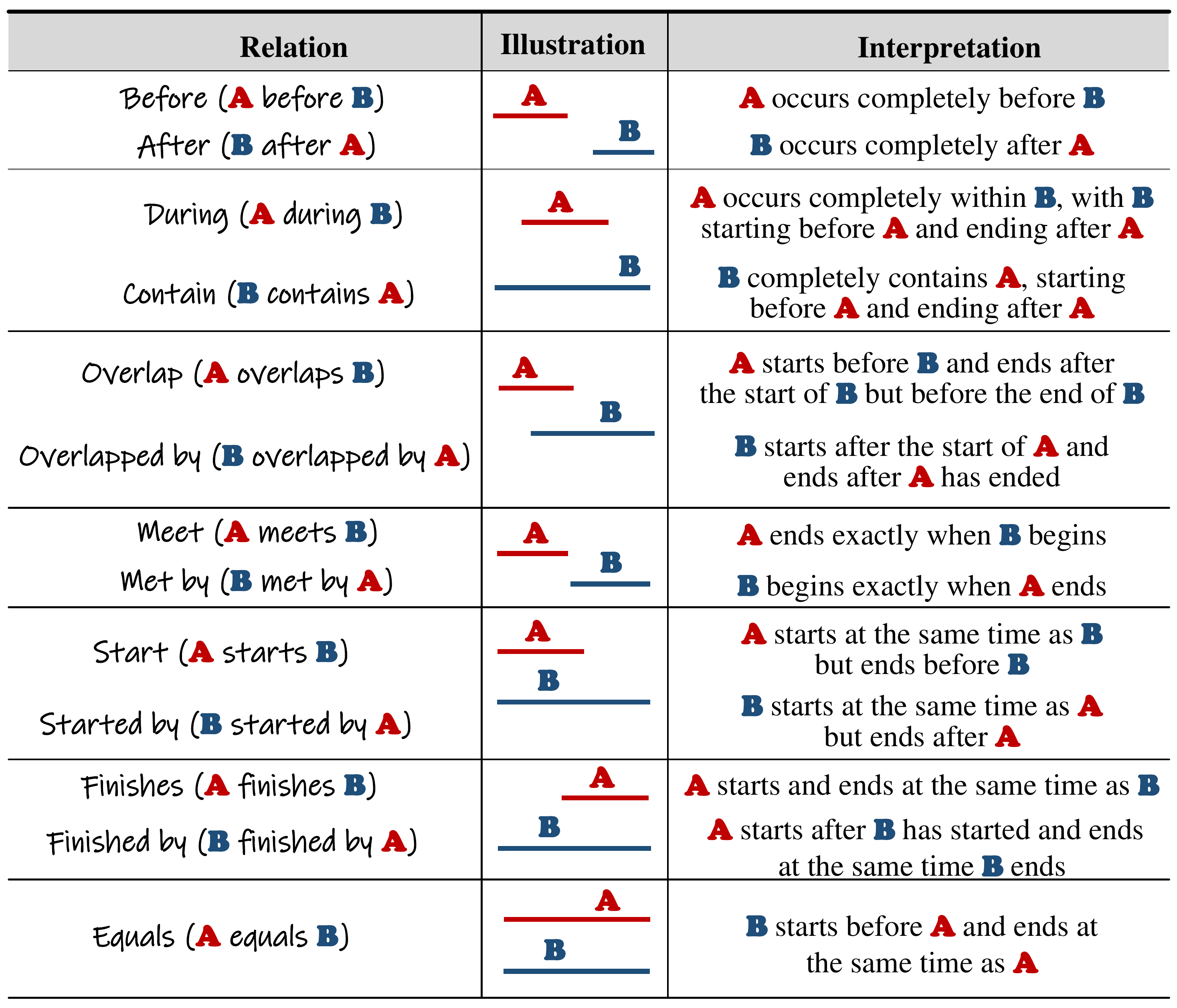}
	\caption{Illustration of 13 base temporal relations defined in Allen's interval algebra~\citep{allen1983maintaining}.} \label{fig:temrelation}
\end{figure}

\noindent\textbf{Temporal relations:} These relations, while not always directly visual, enrich visual knowledge by marking the sequence or timing of events and transformations within a visual scene over time. For example, temporal relations can  describe the progression of actions, such as ``before,'' ``after,'' and ``during,'' which are instrumental in comprehending the dynamics of environments and activities. See Fig.~\ref{fig:temrelation} for the 13 base temporal relations defined by~\cite{allen1983maintaining}.

\begin{figure}[tbh]
	\centering
	\includegraphics[width=0.7\linewidth]{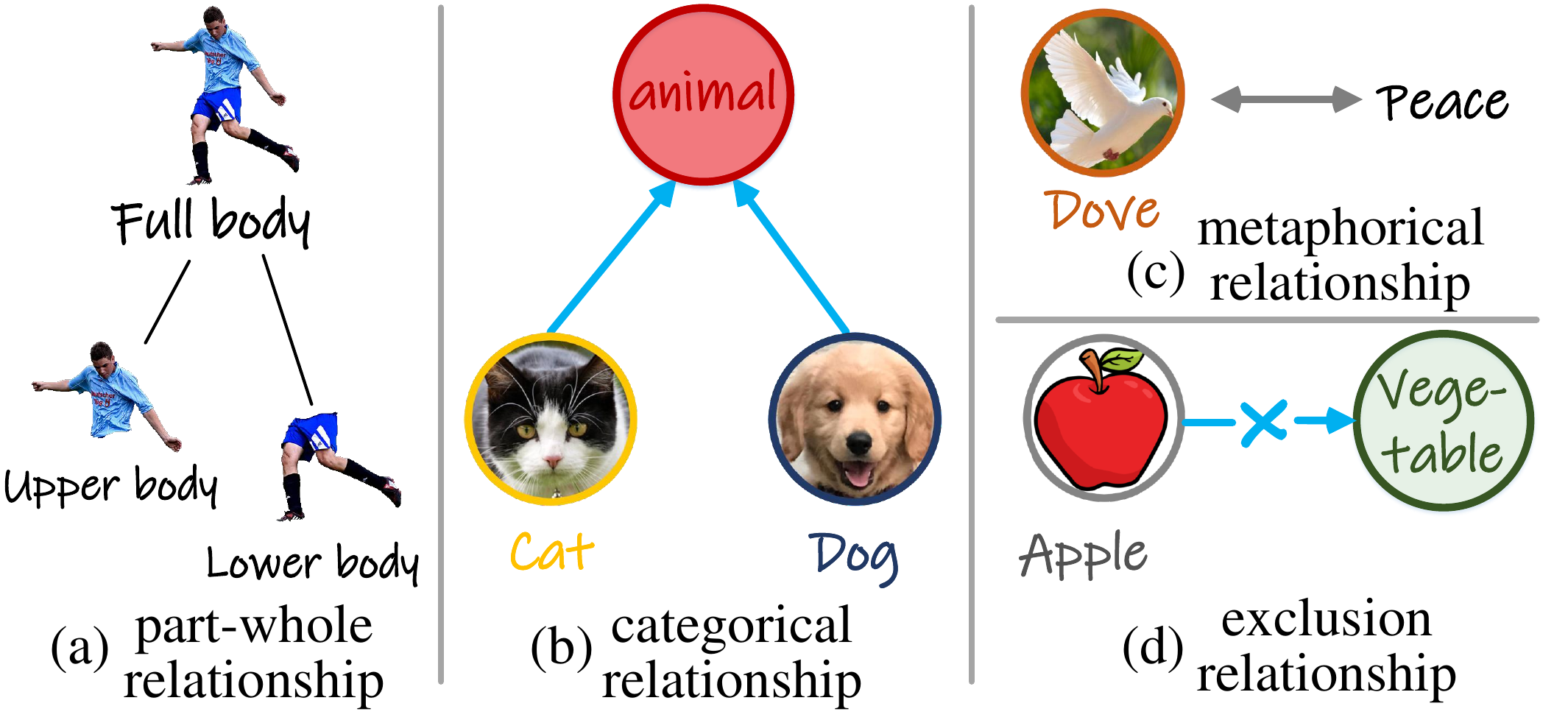}
	\caption{Illustration of semantic relations.} \label{fig:semrelation}
\end{figure}

\noindent\textbf{Semantic relations:} These relations specify the connections between objects or concepts grounded in their meanings or significances, hence enhancing our grasp of meanings, part-whole relationships, similarities, differences, inclusion-exclusion criteria, and semantic dependencies within visual information. For example, the part-whole relationships help us dissect visual concepts into their constituent parts or components, such as an apple can be broken down into its kernels, flesh, rind, and pedicel; human body can be broken down into different parts (Fig.~\ref{fig:semrelation}(a)). Each of these sub-concepts maintains its own semantic connections both among themselves and with the overarching concept. In a similar vein, the categorical relationships describe the fact that visual concepts can be categorized with others that fall under the same category or superordinate concept; these concepts have semantic relations of similarity and difference with each other and with the category. For instance,  dogs and cats are different types of animals  (Fig.~\ref{fig:semrelation}(b)). Moreover, semantic relationships involve more abstract associations, like the metaphorical relationship between ``dove'' and ``peace''  (Fig.~\ref{fig:semrelation}(c)). Furthermore, 
semantic relationships enable the inclusion or exclusion of visual concepts based on specific criteria or rules, such as whether they belong to a certain domain or context. For example, apples and oranges are classified as fruits, not vegetables, based on certain distinguishing factors  (Fig.~\ref{fig:semrelation}(d)). 

\begin{figure}[tbh]
	\centering
	\includegraphics[width=0.7\linewidth]{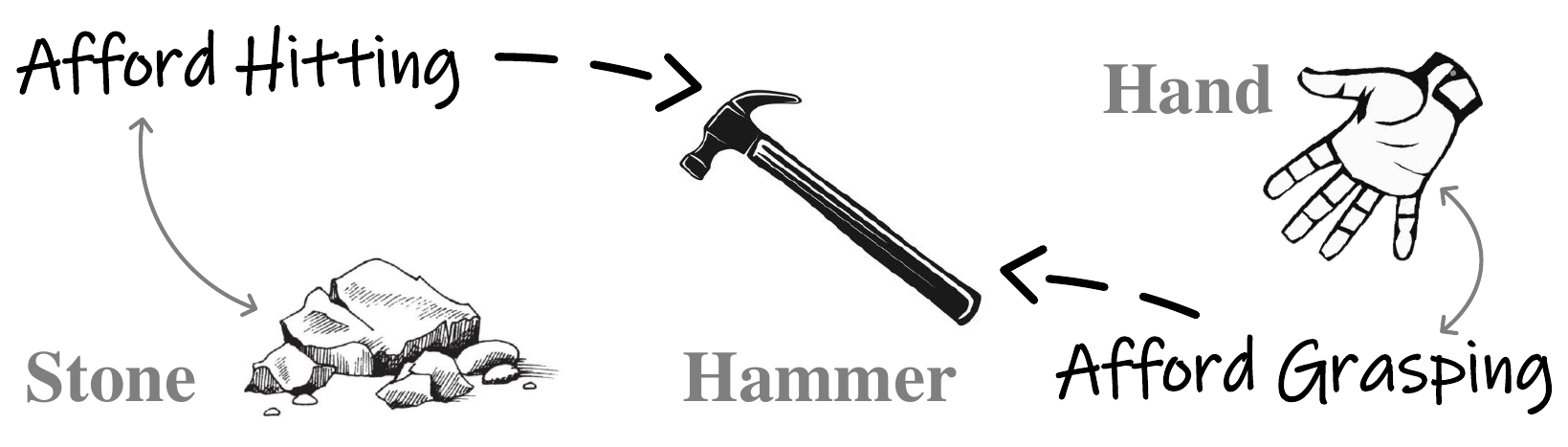}
	\caption{Illustration of functional relations.} \label{fig:funrelation}
\end{figure}

\noindent\textbf{Functional relations:} These relations explicate the interactions among objects based on their physical properties or affordances, hence facilitating$_{\!}$ our$_{\!}$ understanding$_{\!}$ of$_{\!}$ purpose,$_{\!}$ utility,$_{\!}$ effect,$_{\!}$ cause,$_{\!}$ and action-relevant structures. For example, a knife can cut a bread, a chair can support a human, a pen can write on a paper, and so on.  Functional relations establish the causal links between behavior and its environmental antecedents (stimuli) and consequences (reinforcers or punishers). For example, if a child learns that pressing a button produces a sound (antecedent), he or she may press the button more often (behavior) to hear the sound more frequently (consequence). Functional relations are 
are foundational for reasoning and problem-solving, as they allow us to infer new facts or actions from existing facts or actions. By identifying the functional relations between problem behaviors and their environmental variables, one can design interventions that either change the antecedents or consequences of problem behaviors or teach alternative behaviors that serve the same function. For example, we can use functional relations to infer that if a knife can cut a bread, then it can also cut a cheese. We can also use functional relations to infer that if we want to broke stones into pieces, then we need a hammer (Fig.~\ref{fig:funrelation}). Functional relations also help us generate explanations or justifications for facts or actions. For example, we can use functional relations to explain why we use a knife to cut a bread or why we sit on a chair.

\begin{figure}[tbh]
	\centering
	\includegraphics[width=0.8\linewidth]{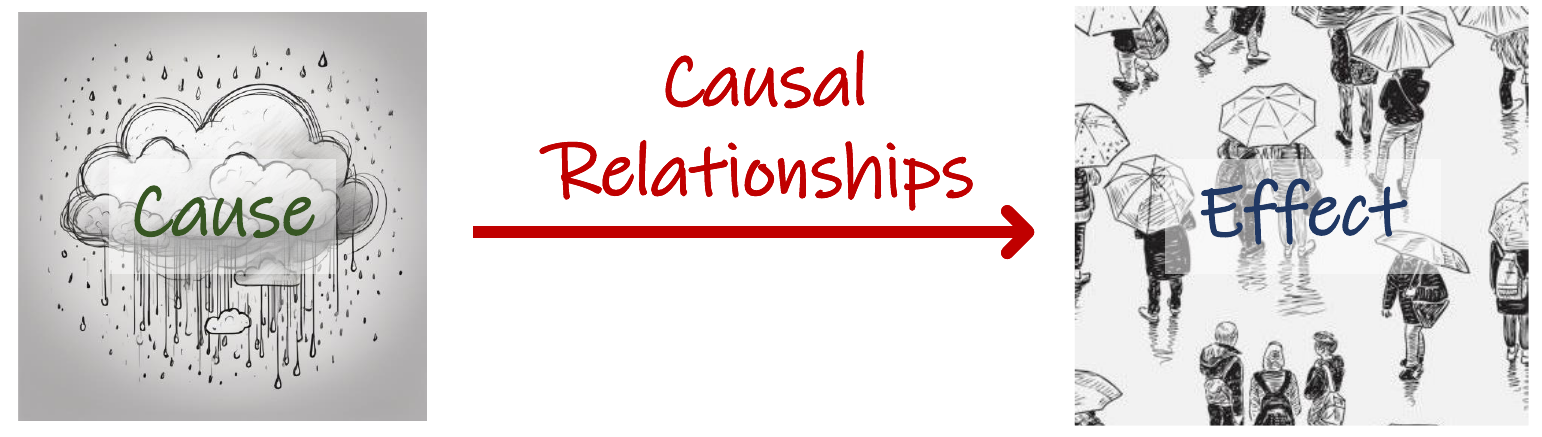}
	\caption{Illustration of causal relations.} \label{fig:caurelation}
\end{figure}

\noindent\textbf{Causal relations:} These relations identify  the cause-and-effect connections between visual elements, which are essential for interpreting how and why changes occur in the visual scene, such as understanding that rain causes the street to be wet (Fig.~\ref{fig:caurelation}). Causal relations also enable predictive reasoning about the outcomes of actions and events within a visual context.

%\noindent\textbf{Comparative and quantitative relations:} These relates are about the comparison of objects and the assessment of their quantitative aspects, facilitating a detailed analysis of numerical attributes and dimension (such as size, quantity, distance, and degree) and providing a quantitative dimension to visual knowledge. Comparative relations include concepts like ``larger,'' ``smaller,'' ``more,'' and ``less,'' while quantitative relations allow for the assessment of numerical attributes and measurements.

Modeling these visual relations is fundamental to the fabric of visual knowledge theory, as it enables AI systems to process and decipher visual information in a structured and meaningful manner. By categorizing and analyzing these relations, researchers can develop more sophisticated artificial models for visual perception, enhancing our ability to replicate human-like understanding and reasoning in machines.

%imagine you are a Cognitive psychologist

%write a long and academic article about functional relation. here is the guideline ""

\subsubsection{Visual operation}\label{sec:223}
The term ``visual operation'' in visual knowledge theory denotes transformations over visual concepts or objects in space or time, such as composition, decomposition, replacement, combination, deformation, motion, comparison, destruction, restoration, and prediction. Visual concepts are the key elements of visual knowledge, enabling us to recognize, categorize, and name the entities we observe in our environment. Furthermore, visual relations enhance our understanding of the  interconnectedness and functionalities of these entities. Yet, as illuminated by cognitive studies from scholars such as~\citep{Margolis1999-MARCCR,carey2000origin,nersessian2010creating,Margolis2015-MARTCM-7,sep-cognitive-science}, visual concepts are subject to manipulation through cognitive processes that, for example, transform them in space or time, alter their components or characteristics, and facilitate various operations over these concepts or objects. These operations are instrumental in augmenting our capacity to comprehend the world, fostering innovation, and executing intricate tasks. They embody the dynamic aspect of visual knowledge, showcasing how static images or scenes can be reimagined or restructured through cognitive engagement:

\begin{figure}[tbh]
	\centering
	\includegraphics[width=0.6\linewidth]{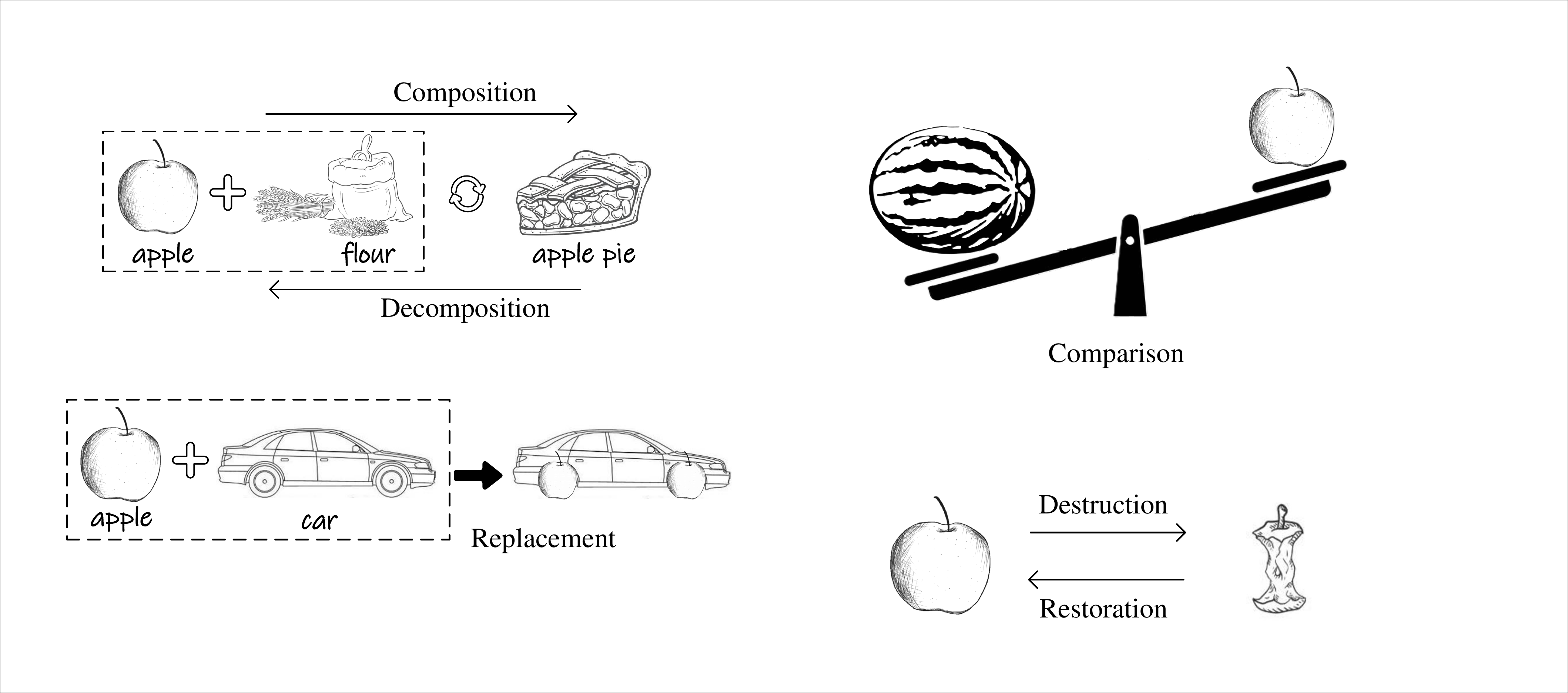}
	\caption{Illustration of the composition and decomposition operations. } \label{fig:ComDe}
\end{figure}

\noindent\textbf{Composition and decomposition:} Composition involves assembling multiple visual elements to form a new object or concept, whereas decomposition refers to breaking down an object into its constituent elements. These operations are crucial for understanding complex systems and structures by analyzing their parts and how they fit together. Moreover, they are essential for the generation of innovative and creative  concepts or objects. For example, through thoughtful arrangement of an apple with other objects (\textit{e.g.}, flour), we can create new inventions (\textit{e.g.}, an apple pie; Fig.~\ref{fig:ComDe}).

\begin{figure}[tbh]
	\centering
	\includegraphics[width=0.7\linewidth]{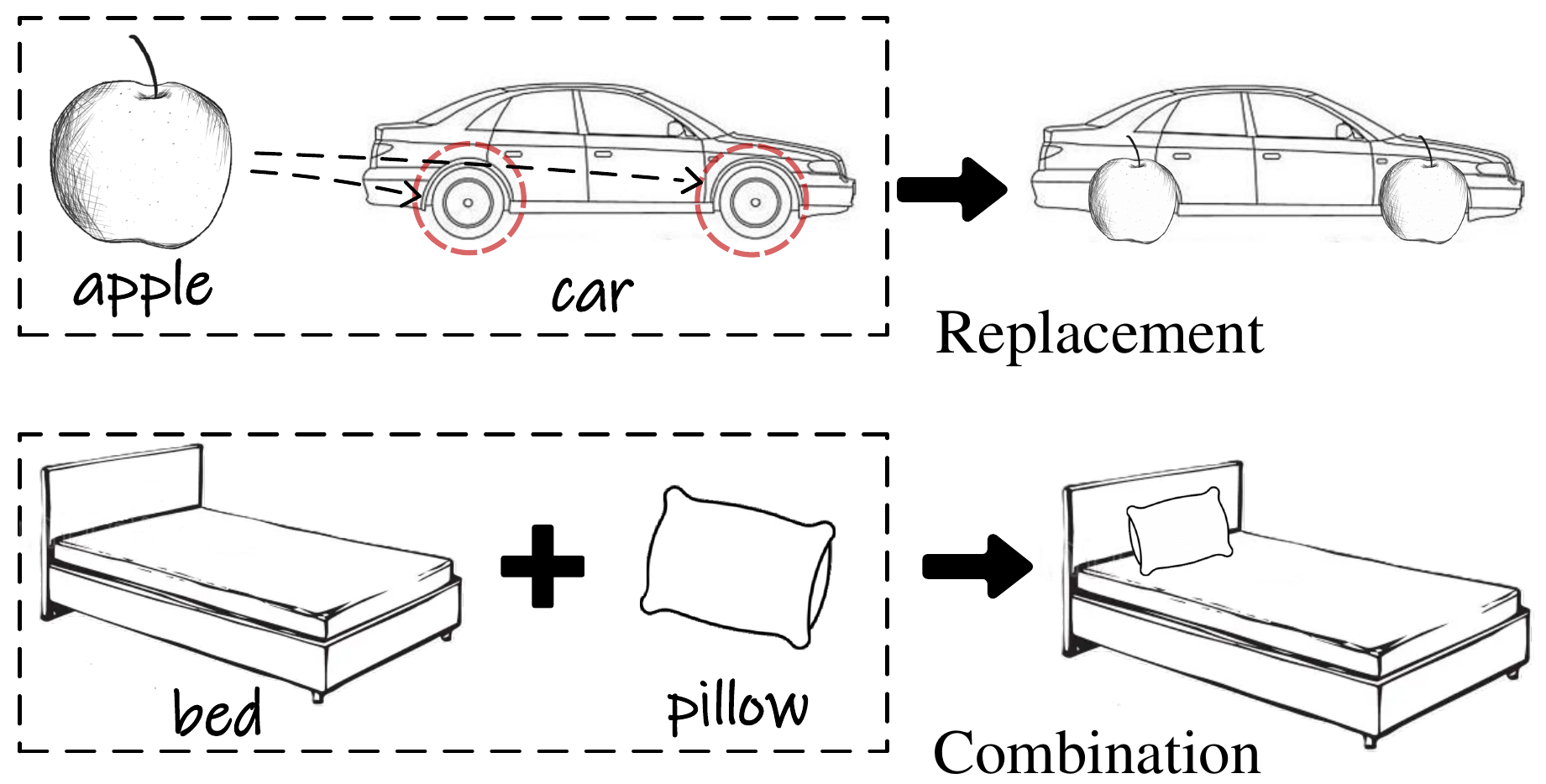}
	\caption{Illustration of the replacement and combination operations. } \label{fig:Rep}
\end{figure}

\noindent\textbf{Replacement and combination:} Replacement entails substitution of one visual element for another, whereas combination pertains to merging distinct elements to forge a new entity. These operations are fundamental to creative thinking and problem-solving, allowing for the exploration of alternative configurations and solutions. They also enhance our understanding of the functionality of objects by allowing for imaginative scenarios, such as replacing the wheels of a car with apples (Fig.~\ref{fig:Rep}).

\begin{figure}[tbh]
	\centering
	\includegraphics[width=0.5\linewidth]{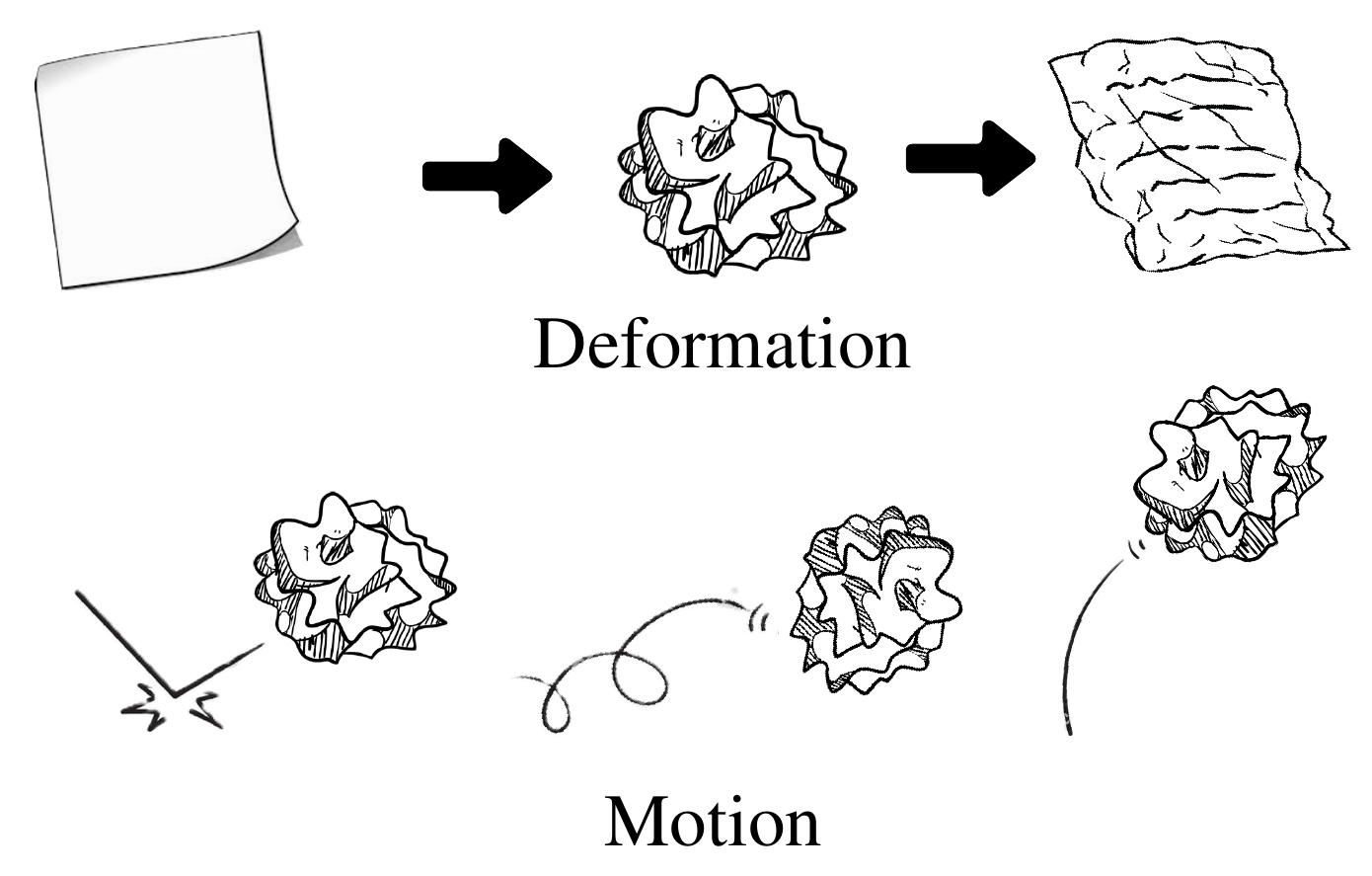}
	\caption{Illustration of the deformation and motion operations. } \label{fig:def}
\end{figure}

\noindent\textbf{Deformation and motion:} Deformation refers to altering the shape or structure of an object, whereas motion involves changing its position over time. Understanding these operations is vital for comprehending both the intrinsic and extrinsic properties of objects, interpreting various physical and biological processes, as well as for crafting animations and simulations that replicate real-world phenomena.  Examples include manipulating an apple by scaling, rotating, or translating it in space; or modifying the motion of a falling apple by speeding up, slowing down, reversing, looping, or interpolating its moving trajectory (Fig.~\ref{fig:def}).

\begin{figure}[tbh]
	\centering
	\includegraphics[width=0.5\linewidth]{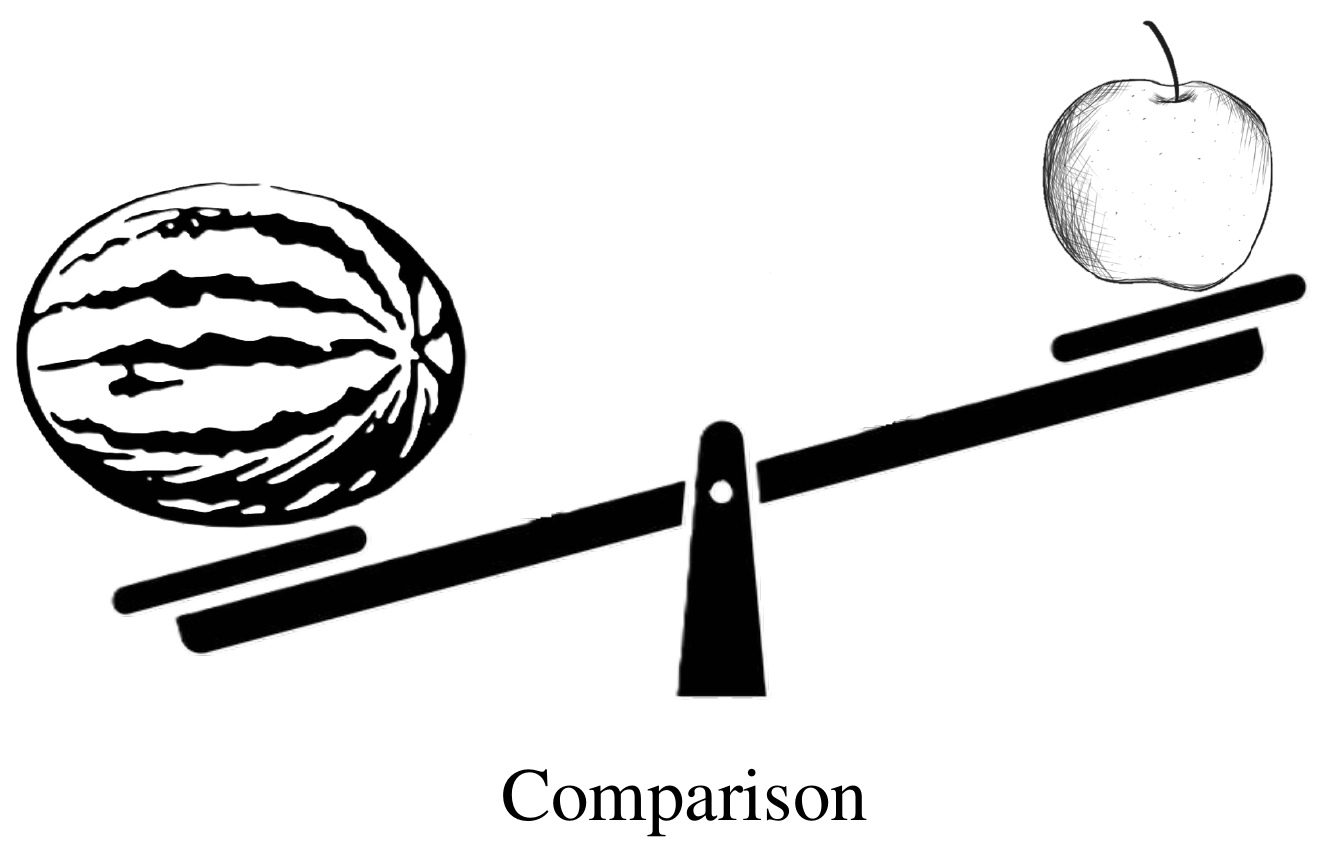}
	\caption{Illustration of the comparison operation. } \label{fig:comp}
\end{figure}

\noindent\textbf{Comparison:} This operation entails evaluating similarities and differences between visual elements, aiding in classification, and decision-making processes. For example, we can compare an apple with other apples or objects in terms of size, weight, \textit{etc} (Fig.~\ref{fig:comp}). Comparison is essential for discerning patterns, making judgments, and learning from visual experiences.

\begin{figure}[tbh]
	\centering
	\includegraphics[width=0.5\linewidth]{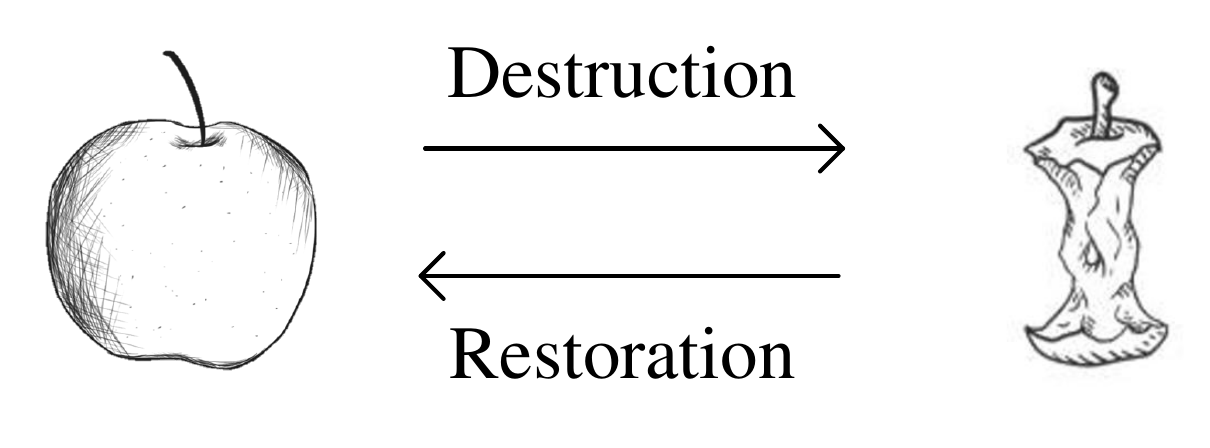}
	\caption{Illustration of the destruction and restoration operations. } \label{fig:Des}
\end{figure}

\noindent\textbf{Destruction and restoration:} Destruction involves the removal or breakdown of visual elements, and restoration focuses on repairing or returning them to their original state (Fig.~\ref{fig:Des}). These operations can be applied in various contexts, from understanding natural disasters and their aftermath to conservation efforts in art and historical preservation.

\begin{figure}[tbh]
	\centering
	\includegraphics[width=0.6\linewidth]{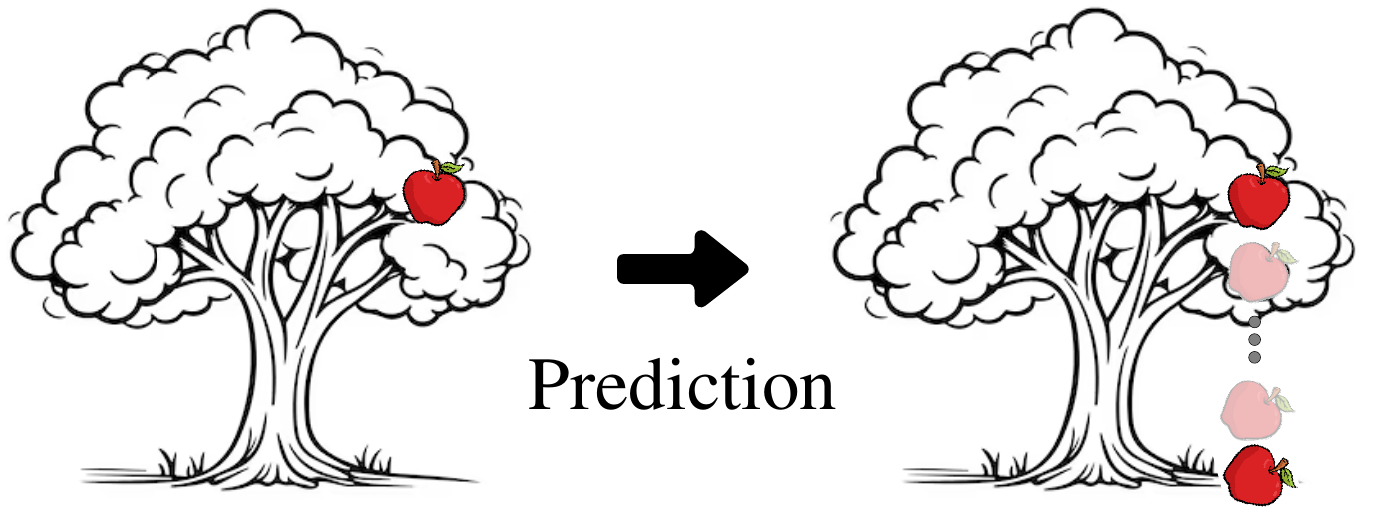}
	\caption{Illustration of the prediction operation. } \label{fig:pre}
\end{figure}

\noindent\textbf{Prediction:} This operation involves projecting future states or changes of visual elements based on current or past information (Fig.~\ref{fig:pre}). This operation is crucial for planning, forecasting, and anticipating outcomes of actions and events.

Through these operations, visual knowledge theory provides a framework for comprehending how visual information can be dynamically manipulated and utilized. These operations underscore the versatility and power of visual knowledge, illustrating its pivotal role in enhancing our ability to interact with, modify, and make predictions about the visual world, as well as its vast potential for application across various domains.

\subsubsection{Visual reasoning}\label{sec:224}

%Visual reasoning refers to the cognitive process of deriving logical conclusions, making decisions, or solving problems based on visual information. It involves interpreting, analyzing, and integrating visual data with existing knowledge to reason about the world, make predictions, or understand complex scenarios. Visual reasoning encompasses the ability to use visual concepts, relations, and operations to think about and reason with visual phenomena.

\begin{figure}[tbh]
	\centering
	\includegraphics[width=\linewidth]{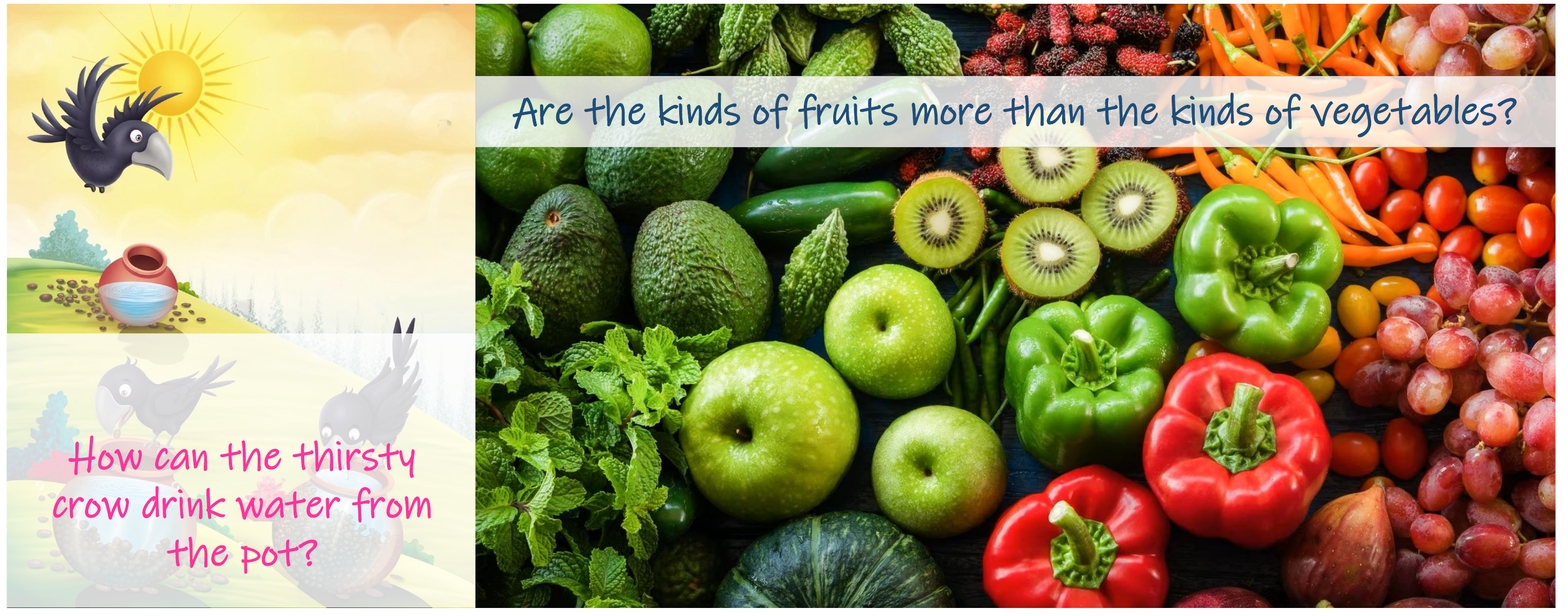}
	\caption{Two examples for visual reasoning.} \label{fig:visual_reasoning}
\end{figure}

The term ``visual reasoning'' in visual knowledge theory denotes the process of applying the knowledge gained from visual concepts, relations, and operations to interpret visual data, solve problems, and make informed decisions (Fig.~\ref{fig:visual_reasoning}). This intricate process usually entails a series of methodical operations on visual concepts and relations, aimed at deriving valid and sound conclusions from what they observe visually and already know (common sense and knowledge). %premises or evidence

In short, visual concept (Sec.~\ref{sec:221}) is about the identification and categorization of visual items; visual relation (Sec.~\ref{sec:222}) is about understanding the connections or associations between these items; visual operation (Sec.~\ref{sec:223}) is about the processes applied to manipulate or analyze visual items; and visual reasoning is about the process (Sec.~\ref{sec:224}) that uses visual concepts, relations, and operations to solve problems, make decisions, or derive sound conclusions from visual information.

\section{$_{\!\!\!}$Visual$_{\!}$ knowledge$_{\!}$ in$_{\!}$ the$_{\!}$ pre$_{\!}$ big$_{\!}$ model$_{\!\!}$ era: retrospect}\label{sec:re}

In this section, we provide an overview of recent research relevant to visual knowledge, along the four key components of visual knowledge outlined in Sec.\!~\ref{sec:ob}, namely visual concept, visual relation, visual operation, and visual reasoning.

\subsection{Visual knowledge: visual concept}

The idea of representing visual concepts by \textit{prototype} and \textit{scope} (Sec.~\ref{sec:221}) has been explored in a few fundamental computer vision tasks, namely image classification and segmentation. For instance, prototype-based networks~\citep{DBLP:conf/nips/SnellSZ17}, non-parametric neural classifiers~\citep{zhou2022rethinking}, and nearest centroids based neural classifiers~\citep{DBLP:conf/iclr/WangHZL23} were developed, where each class is represented by one or a few prototypes, and new observations are classified by their proximity to the class prototypes. Despite their impressive performance in few-shot and general settings, these approaches fail to capture the scope of each class/prototype. For more comprehensive modeling of the underlying data distribution, deep generative classifiers~\citep{DBLP:conf/nips/LiangWM022} were devised to estimate the data density of each visual concept/class as a Gaussian Mixture Model (GMM), where the prototypes and scopes are the estimated GMM's parameters (\textit{i.e.}, the mean vectors and covariance matrices). Its remarkable
results on both closed-set and open-world scenarios evidence the power of {prototype} and {scope} based visual concept representation.

\subsection{Visual knowledge: visual relation}
As discussed in~Sec.~\ref{sec:222}, visual concepts can be related to each other in different ways, resulting in various types of visual relations, such as geometric, temporal, semantic, functional, and causal relations.%, comparative, and quantitative relations.

Geometric relations describe how objects are arranged and transformed in space, including their position, orientation, size and shape. Capsule network \citep{DBLP:conf/nips/SabourFH17} is a landmark effort for modeling the geometric relations among visual elements. Basically, a capsule is a collection of neurons whose activity vector represents the probability and the pose of a visual concept or object. The pose is a set of parameters describing the spatial relation and transformation of the visual concept, such as its location, rotation, scale and reflection. Though theoretical impressive, capsule networks are  less practical for real-world applications, suggesting the great challenge of visual geometric relation modeling.

%Empirical analysis shows that each dimension of the capsule vector encodes different spatial pose information of  MNIST handwritten digit. However, capsule network does not show prominent performance on more challenging tasks.

Semantic relations specify how objects are related to each other in terms of their meanings. A set of recent efforts are devoted to explore semantic relations in the context of visual understanding~\citep{[xu]li2022deep,li2023semantic} and human parsing~\citep{wang2019learning,wang2020hierarchical}. For instance, \cite{[xu]li2022deep} proposed a neural parser capable of generating structured, pixel-wise descriptions of visual observations in terms of a semantic concept hierarchy. This structured visual parser make an explicit use of the composition and decomposition dependencies among semantic concepts as additional regularization terms during network training. For example, an observation that is likely to be a cat should have a low probability of being any vehicle subcategory. However, it is important to note that the semantic relations encoded in the class hierarchy are predefined, rather than automatically learned. This suggests learning visual semantic relations is a challenging problem that requires much further work.

Temporal relations explicate the sequential or chronological order of events and actions as they occur over time within visual data. Temporal relations are mainly studied in the fields of action recognition and video object detection. For instance, in the field of action recognition, Something-Something dataset~\citep{goyal2017something} requires fine-grained motion distinctions and temporal modeling to distinguish interactions like \textit{picking something up} and \textit{putting something down}, providing a good testbed for temporal relation understanding. Video object detection is also a classic computer vision task with the target of classifying, segmenting, and tracking object instances in video sequences~\citep{russakovsky2015imagenet,zhou2022survey}.

Functional relations refer to the actions that objects enable or support. Various tasks in computer vision study functional relation of visual concepts, such as human-object interaction (HOI) detection, and  affordance estimation (a.k.a functional recognition). HOI detection~\citep{gupta2009observing,zhou2021cascaded,li2023neural} aims to locate and identify the relationships between human and objects in visual scenes, like $<\!\!\textit{girl}, \textit{eat}, \textit{apple}\!\!>$, while affordance estimation~\citep{stark1991achieving,li2023beyond} is to predict typical action-object affordances from visual information, such as \textit{eatable}, \textit{openable}.

Another related task to visual relation understanding is scene graph generation (SGG)~\citep{johnson2015image,krishna2017visual},  which is to generate a visually-grounded graph as an explicit structural description of a visual scene. The nodes in a scene graph represent the objects and the edges represent the relationships between the objects (including spatial, part-whole, and interaction relationships). Each relationship between two objects is denoted as a triplet of $<\!\!\texttt{subject}, \texttt{PREDICATE}, \texttt{object}\!\!>$, \textit{i.e.}, $<\!\!\texttt{boy}, \texttt{RIDE},$ $\texttt{car}\!\!>$, $<\!\!\texttt{car}, \texttt{HAS}, \texttt{wheel}\!\!>$, $<\!\!\texttt{car}, \texttt{NEAR}, \texttt{building}\!\!>$. Although SGG, to some extent,  encompasses the three types of visual relations: geometric, semantic, and functional relations,  in one single task, the covered relations are still sparse, for example, the functional relations are typically human-centric actions, less considering object-centric affordances. Moreover, SGG requires costly human efforts for visual relation annotation.

Causal relations delineate how events, actions, or objects within a visual context can directly influence or result in one another. 
Recent years witnessed unprecedented achievements of the deep learning technique across various domains, which, however, relies heavily upon fitting the data distributions. It is apparent that such technique tends to only learn correlation-based patterns (statistical dependencies) rather than the essential causal relationships  from data and thus easily collapses into data bias and has limited generalization  ability~\citep{hendrycks2018benchmarking}. In response to this challenge, a set of recent efforts have shifted towards uncovering the causality embedded the visual data. Basically, these efforts investigate causal reasoning~\citep{pearl2009causality,scholkopf2021toward} within the deep learning framework to extract causal representations from visual data. Leveraging causality-guided visual representations, they achieved notable performance in tasks such as visual recognition~\citep{yue2021counterfactual} and visual question answering (VQA)~\citep{wang2020visual}. They also showed that they are capable of automatically discovering causal dependencies among environmental and object variables from videos~\citep{li2020causal}, and improving the interpretability~\citep{shi2021temporal} and out-of-distribution generalization ability~\citep{christiansen2021causal} of deep learning models.

\subsection{Visual knowledge: visual operation}
% concept的改变 （猫和动物改变， 保持concept改scope）
% 几何和语义关系的改变（AIGC、style transfer）

As discussed in Sec.~\ref{sec:223}, visual operation refers to the manipulation of over visual concepts or objects in space or time. A closely related research field is customized visual content generation, aiming at generating
creative contents for a target novel concept guided by textual descriptions. The textual descriptions serve as a feasible and flexible tool for specifying editing intensions, allowing for diverse visual operations such as replacement and combination. The task of generating realistic images from natural language descriptions -- text-to-image synthesis -- has been a research focus for years. Various deep generative models
have been established for this task, such as GANs~\citep{reed2016generative}, VAEs~\citep{ramesh2021zero}, and Autoregressive Models~\citep{DBLP:journals/corr/abs-2204-06125}. 
Recently, diffusion models have demonstrated remarkable ability in generating text-aligned images with high fidelity~\citep{rombach2022high,saharia2022photorealistic}. 
However, they encounter difficulties in performing customized generation for novel concepts, such as a specific animal or object which only appear in single testing image. Various customized visual content generation approaches have been developed to address this challenge~\citep{gal2022image,ruiz2023dreambooth}. They lift pre-trained text-to-image synthesis models to synthesize novel scenes of target concepts (typically represented by one or a few user-provided reference images) under natural language instruction. They have proven capable of generating not only creative static images~\citep{gal2022image,ruiz2023dreambooth} but also temporally coherent videos~\citep{xing2023make} that adhere to the guidance intentions. Though impressive, they may struggle for complicated visual operations (like decomposition, destruction, and restoration).

Novel view synthesis, \textit{i.e.}, synthesizing new images of the same object or scene from arbitrary viewpoints given single or multiple inputs of the object/scene, is relevant to two visual operations: deformation and motion. Building on the concept of novel view synthesis, it becomes imperative to discuss the intricacies involved in simulating deformation and motion for objects within these synthesized viewpoints. This requires a deep understanding of the spatial and temporal aspects of objects, allowing for the generation of images that not only look realistic but also behave in ways that are consistent with the physical world. Techniques such as 3D modeling and neural radiance fields (NeRFs)~\citep{mildenhall2020nerf} have been instrumental in advancing this area. NeRFs, in particular, have shown great promise in creating detailed and continuous volumetric scenes, enabling smooth transitions and realistic deformations across different views~\citep{pumarola2021d}. However, they require significant computational resources for both training and inference. Recently, 3D Gaussian splatting~\citep{kerbl20233d,chen2024survey}, which represents a 3D scene with millions of 3D Gaussians, has demonstrated remarkable ability in real-time rendering. By introducing additional spatial-temporal modules~\citep{yang2023real} or Gaussian properties~\citep{luiten2023dynamic}, they can model the dynamics and deformation in a given scene.

Regarding prediction, the visual operation of forecasting future states, actions, or events from visual data, has garnered significant interest, with numerous pertinent tasks in computer vision. Some representative tasks are: 
\begin{itemize}%[leftmargin=*]
   \setlength{\itemsep}{0pt}
   \setlength{\parsep}{-2pt}
   \setlength{\parskip}{-0pt}
   \setlength{\leftmargin}{-10pt}
   \vspace{-4pt}
   \item Human trajectory prediction: estimate the future paths of people in various settings, such as pedestrians on sidewalks or shoppers in malls, considering social behaviors and surroundings~\citep{alahi2016social}. 
   \item Future frame prediction: generate future frames of a video sequence to accurately depict the continuation of the observed scene~\citep{mathieu2016deep}.
   \item Action prediction: anticipate the future actions of subjects in a video, such as predicting an athlete's next move in sports or a driver's behavior in traffic~\citep{ryoo2011human}.
   \item Physical interaction forecasting: predict the outcome of physical interactions between objects, such as forecasting the motion of objects after a collision~\citep{watters2017visual}.
   \item  Accident anticipation: identify potentially hazardous situations before they occur, such as anticipating vehicle collisions or industrial accidents from surveillance footage~\citep{suzuki2018anticipating}.
   \vspace{-4pt}
\end{itemize}

\subsection{Visual knowledge: visual reasoning}
Visual reasoning is the process of applying visual concept, visual relation, and visual operation, which are commonalities among various tasks, to draw valid and sound conclusions from premises or evidence~\citep{yunhe1996}. We can use functional relations to infer like ``If $A$ can cut $B$, then $B$ is soft'' or ``If $A$ can support $B$, then $A$ is stable'' (Sec.~\ref{sec:223}). Reasoning is a fundamental cognitive process that enables humans to make sense of the world. Human reasoning is a complex and multifaceted phenomenon that can be categorized into different types, such as deductive, inductive, abductive, and analogical  reasoning, depending on the nature and strength of the arguments involved. Machine reasoning is a field of AI that complements the field of machine learning by aiming to performing automated reasoning. This is done by way of uniting known (yet possibly incomplete) information with background knowledge and making inferences regarding unknown or uncertain information, with the aid of efforts at many different disciplines, such as cognitive neuroscience, psychology, linguistic, and logic~\citep{bottou2014machine}. 

Early explorations of automated reasoning systems primarily adopted two approaches: connectionism and symbolism. In the 1940s, McCulloch and Pitts proposed the first simplified neuronal model~\citep{mcculloch1943logical}, establishing the foundation for research in neural networks and connectionism. From the perspective of connectionism, reasoning is the result or derivation of multiple, interconnected, simple processing devices, one major example being neural networks. The main motivation behind connectionism comes from cognitive neuroscience, since the human neural circuitry is clearly capable of storing and retrieving knowledge organized in short-and long-term memory, by continuously analyzing and processing new, complex information, and reasoning upon it. While connectionist models, particularly neural networks, are adept at capturing statistical patterns from data, they are bounded to computational resource and data availability at that time. Hence achieving the nuanced reasoning for human-like inferences remains intricate. This challenge prompted the exploration of symbolism, which has its roots in the study of logic and philosophy and became the dominate approach to Good-Old-Fashioned AI from the middle 1950s to the late 1980s. Basically, symbolic approaches posit that symbols representing worldly objects and concepts form the foundational building blocks of human intelligence. They see reasoning as the capability of deriving additional information from that already encoded in a collection of given symbols, by performing elaboration and manipulation on the given structured symbolic representations. They typically conduct reasoning by applying a series of restrict rules and formal logic operations to manipulate the discrete symbols in a rigorous and precise way. The rules defined how symbols could be manipulated to draw conclusions or make decisions. For example, a symbolic rule might deduce that ``Socrates is mortal'' from the premises that ``All men are mortal'' and ``Socrates is a man.'' Consequently, symbolic approaches are powerful for problems with well-defined rules and discrete values, offering transparent explanations for their reasoning processes and allowing to check the validity or satisfiability of their logical steps. However, they are intolerant of ambiguous and noisy data, making them less suitable for many real-world applications due to the difficulty in defining the hard rules and the high degree of uncertainty involved. Probabilistic reasoning techniques, such as Bayesian networks, Markov decision processes, and stochastic models, use probability theory to represent and manipulate uncertainty, and can provide probabilistic estimates for the outcomes of reasoning. While these techniques can empower symbolic methods with the ability to deal with uncertainty, the intrinsic limitations of symbolic approaches, namely the lack of true learning and the reliance of hand-crafted rules -- remain significant obstacles to their application in practical contexts.
%Probabilistic reasoning is particularly effective in situations where data is incomplete, noisy, or ambiguous, allowing for decisions to be made with a known degree of confidence.

%Historically, there has always been a dichotomy between symbolic and subsymbolic (often named connectionist) frameworks to model reasoning [4].

%Probabilistic Reasoning: These models handle uncertainty by using probabilities to represent the likelihood of various outcomes. Bayesian networks, in particular, are effective for reasoning under uncertainty, allowing AI systems to make inferences and decisions even when information is incomplete or uncertain. 

Recently, the emergence of large-scale datasets coupled with significant advancements in computational resources has sparked a resurgence in connectionism, particularly rejuvenating interest in neural network algorithms that are decades old. Yet, despite their prowess in pattern recognition and predictive modeling, DNNs struggle with reasoning tasks that necessitate explicit manipulation of symbols. Specifically, while DNNs excel at learning subsymbolics (\textit{i.e.}, continuous embedding vectors), their architecture is not inherently suited for discrete symbolic operations that reasoning often entails.
Moreover, DNNs typically learn from data in an inductive manner, which contrasts with the deductive procedure prevalent in reasoning, where logical deductions are drawn from explicit, pre-established rules and knowledge bases. It is not straightforward to integrate domain-specific knowledge into DNNs for explicit reasoning. In addition, the decision-making process of DNNs is often obscured, making it challenging to comprehend how particular conclusions are reached. This opacity is particularly problematic in decision-critical applications such as autonomous driving.  On the other hand, symbolic approaches, though far less trainable, are excellent at principled judgements (such as deductive reasoning), and exhibit inherently high explainability (as they operates on clear, logical principles that can be easily traced and understood). In light of the challenges faced by DNNs in explicit reasoning and considering the complementary nature of connectionist and symbolic methodologies, a pioneering research domain, termed neuro-symbolic computing (NeSy), has gained prominence~\citep{DBLP:journals/flap/GarcezGLSST19}. NeSy essentially pursues the principled integration of the two foundational paradigms in AI, providing a new framework of more powerful, transparent, and robust reasoning~\citep{wang2022towards}. 

Traditionally, tasks related to visual reasoning are typically visual question-answering (VQA) and visual semantic parsing. VQA, \textit{i.e.}, answering questions based on visual content, requires comprehensive understanding and reasoning over both the visual and linguistic modalities. \cite{andreas2016neural} introduced a NeSy based VQA system which interprets questions as executable programs composed of learnable neural modules that can be directly applied to
images. A module is typically implemented by the neural attention operation and corresponds to a certain atomic reasoning step, such as recognizing objects, classifying colors, \textit{etc}. This pioneering work stimulated many subsequent studies that explore NeSy for approaching VQA~\citep{yi2018neural,vedantam2019probabilistic,amizadeh2020neuro}. Visual semantic parsing seeks for a holistic explanation of visual observation in terms of a class hierarchy. The class hierarchy, pre-given as a knowledge base, encapsulates the symbolic relations among semantic concepts. \cite{li2023logicseg}~devise a powerful NeSy based visual semantic parser through end-to-end embedding symbolic logic into both the network's training and inference stages. Some other relevant tasks include visual abductive reasoning~\citep{liang2022visual} and visual commonsense reasoning~\citep{zellers2019vcr}.

More recently, benefiting from the impressive emergent capabilities of large language models (LLM), some efforts explore LLMs for solving sophisticated visual reasoning tasks. VisProg~\citep{gupta2023visual} represents a pioneering effort in this domain; it uses LLMs to decompose visual reasoning tasks (such as ``Is it true that the two images contain a total of six people and two boats?'') into a series of manageable subtasks (such as text parsing, object detection, and counting) and solving them step-by-step. Later, HuggingGPT~\citep{shen2023hugginggpt} leverages LLMs to manage AI models available on the web to solve complicated reasoning tasks. DoraemonGPT~\citep{yang2024doraemongpt} advances this research trend towards solving real-world tasks that involve dynamic observations. It equips LLMs with a symbolic memory for gathering and storing task-relevant information from the dynamic observation, a rich set of extra knowledge sources (\textit{e.g.}, AI tools, search engines, text books, knowledge databases) for reference, and a Monte Carlo Tree Search based planner for efficiently probing the huge solution space. 

\subsection{Discussion}

So far, we have reviewed prior key contributions relevant to the four foundational aspects of visual knowledge. From such review, some key insights can be derived as below:

First, visual knowledge is closely linked to the two fundamental AI paradigms (namely connectionism and symbolism), several research domains (including computer vision, graphics, machine learning, and logic),  a set of fundamental and challenging tasks (including visual recognition, affordance estimation, text-to-image synthesis, novel view synthesis, and future forecasting), and various advanced techniques (including capsule networks, neuro-symbolic computing, and large language models). This underscores the critical significance of visual knowledge, and the necessity of interdisciplinary collaboration to achieve this grand goal.

Second, while our community has indeed achieved progress in certain areas related to visual knowledge, numerous core issues remain challenging and/or underexplored, such as prototype-and scope-based visual concept, causal relation, complex visual operations (\textit{e.g.}, decomposition, destruction, and restoration), and visual reasoning. This highlights the difficulties in creating visual knowledge. This also unveils a primary motivation for proposing visual knowledge -- the lack of a principled framework that offers a unified perspective on different aspects of visual intelligence. 

Third, although LLMs have demonstrated remarkable capabilities in solving intricate problems, they also exacerbate the ``black box'' issue, an innate character of neural network algorithms. With their billions or even trillions of parameters, LLMs pose an insurmountable challenge for anyone attempting to dissect their internal workings. Moreover, LLMs are still yet to perform logical reasoning in the way humans do. They frequently produce plausible-sounding answers that, upon closer examination, reveal a lack of genuine comprehension. Compounding this issue is the opaque nature of LLMs, which obstructs the identification and correction of errors within the reasoning process. In the subsequent section, we will delve into the significance of investigating LLMs within the visual knowledge framework.

\section{$_{\!\!\!}$Visual$_{\!}$ knowledge$_{\!}$ in$_{\!}$ the$_{\!}$ big$_{\!}$ model$_{\!}$ era:$_{\!\!\!}$ prospect}\label{sec:pr}
%According to cognitive psychology, knowledge refers to the mental representation of various types of information that humans learn and experience. Knowledge is formed, stored, updated, and evolved in the human brain through various cognitive processes, such as perception, attention, memory, reasoning, problem-solving, and metacognition. These processes enable humans to encode, organize, retrieve, and apply knowledge in different situations and domains. Such an innate human ability remains challenging for big models to achieve. 

In this section, we first explore how visual knowledge can power big models to bring the level of general intelligence closer to that of humans. We then investigate how big models can boost the development of visual knowledge, given the significant challenges of establishing visual knowledge.

\subsection{Empower$_{\!}$ big$_{\!}$ models$_{\!}$ with$_{\!}$ visual$_{\!}$ knowledge$_{\!\!\!}$}

%achieving remarkable success across a spectrum of domains from natural language processing to complex decision-making systems.  
The advent of large AI models has undeniably marked a new era in AI,  providing unprecedented accuracy and fluency in tasks that were once considered insurmountable for machine. However, despite their astonishing successes, these powerful models are not without their limitations. Among the most critical challenges they face are issues related to transparency, reasoning, and catastrophic forgetting. In the forthcoming discussion, we will show that, despite these substantial obstacles, the integration of visual knowledge into the large models offers a promising avenue for advancement.

%delve deeper into these limitations and show that incorporating visual knowledge into large AI models offers a promising path to overcome these limitations.

%emerging evidence suggests that integrating visual knowledge into large models could offer a promising solution 

One of the most discussed limitations of large models is their lack of transparency. Transparency refers to the degree to which the internal workings and outputs of models can be understood and explained by humans. Due to the sheer volume of parameters, understanding how the big models arrive at a particular conclusion is extremely challenging. The lack of transparency hinders our ability to fully trust and verify the models' decisions, especially in critical applications such as healthcare diagnostics, and autonomous driving, and causes a lot of concerns regarding accountability, bias, fairness, debugging, \textit{etc}. Although there are some network interpretation techniques based upon the analysis of reverse-engineer importance values or sensitivities of inputs, they only produce posteriori explanations for already-trained DNNs. They essentially \textit{approximate} the behavior of DNN by modeling relationships between features and the outputs. Such \emph{post}-\emph{hoc} explanations are problematic and misleading, as they cannot explain what actually makes a DNN arrive at its decisions~\citep{rudin2022interpretable}. Yet, due to the inherent transparent nature of prototype-and scope-based visual concept, the integration of visual knowledge may endow big models with promising \emph{ad}-\emph{hoc} interpretability. A notable evidence is the groundbreaking study by~\cite{DBLP:conf/iclr/WangHZL23}, which introduces Deep Nearest Centroids (DNC), a fully end-to-end, prototypes based neural classifier. Through representing visual concepts as a collection of automatically discovered prototypes (\textit{i.e.}, class sub-centroids), DNC mirrors the experience-/case-based reasoning process that we humans are accustomed to and yields a powerful yet \emph{ad}-\emph{hoc} interpretable framework for large-scale visual recognition. This idea can be further explored for prototype-and scope-based visual concept modeling. By employing such inherently transparent visual concepts as foundational elements, it is naturally anticipated that visual knowledge will enhance the transparency of big models.

While large AI models excel at pattern recognition and generating human-like text or images, they may fail to grasp the underlying logic or truth of the content it produces, hence struggling with tasks requiring an understanding of causality, abstract concepts, or logical inference. For instance, big models may generate plausible-sounding but factually incorrect or nonsensical answers, known as ``hallucination'', reflecting a surface-level mimicry of human output rather than genuine comprehension~\citep{ji2023survey}. The challenge stems from the big models' reliance on statistical patterns or superficial features rather than causation that may not capture the underlying causal relationships or logic. Although a few reasoning strategies such as Chain of Thoughts~\citep{DBLP:conf/nips/Wei0SBIXCLZ22} and Tree of Thoughts~\citep{yao2023tree} are recently developed for boosting large  models' reasoning ability, they are still far from the true reasoning that typically involves managing complex operations over symbolic concepts,  understanding causality, and applying abstract principles to novel situations. Yet, visual knowledge provides an explicit, powerful, and unified framework for the comprehensive modeling of visual conceptions, visual relations (including causality), visual operations, and visual reasoning. As a result, this may bring the reasoning of big models into a brand-new era where the reasoning is powered by both implicit knowledge from big models and explicit knowledge modeled by visual knowledge. In such wise, big models can perform complicated tasks like humans, \textit{e.g.}, reasoning about complex and dynamic scenarios that involve multiple entities and complex relations, solving problems that require series of methodical operations on visual concepts and relations, and applying learned knowledge to fundamentally different problems. Given the recent impressive progress in the integration of symbolic-knowledge based logic reasoning and data-driven neural subsymbolic learning~\citep{li2023logicseg}, we firmly believe that combing big models' implicit knowledge and explicit visual knowledge in a form of multiple knowledge representation~\citep{pan2020multiple,yang2021multiple} is a promising pathway forward.

% leading to difficulties in  or recognizing when extrapolations are invalid.

%Large AI models may not have true understanding of the data they process or the tasks they perform, but rather rely on statistical patterns or superficial features that may not capture the underlying causal relationships or logic. This 

%Limited knowledge: AI models can only learn from the data they are given, and cannot generalize or reason beyond their scope2.

Catastrophic forgetting refers to the tendency of  DNNs to lose their previously learned knowledge when exposed to new data or tasks. 
The root of this issue lies in the way DNNs update their parameters; new learning can overwrite the weights and biases associated with old knowledge, leading to a rapid degradation of performance on tasks the model had previously mastered. Catastrophic forgetting is a fundamental challenge to the vision of creating AI systems that can learn and adapt over time in a manner analogous to human learning. At the heart of catastrophic forgetting lies the difficulty of knowledge trace within big AI models. Knowledge trace, or the ability to identify, follow, and understand the representation and processing of information within a model, is about knowing what the model knows and how it came to know it. In human learning, knowledge trace allows for the accumulation of experiences and the seamless integration of new information with existing knowledge. However, in large AI models, identifying the specific components responsible for particular pieces of knowledge is a daunting task due to the complex network architecture of massive interconnected parameters. Visual knowledge, with its deep root in cognitive psychology, offers large models with a form of knowledge representation that is explicit, structured, persistent, editable, and traceable. This allows to update knowledge outside the big models, enabling more targeted interventions to prevent catastrophic forgetting. Moreover, augmented with visual knowledge, the big models can create more durable and retrievable memory to enhance recall and understanding, like humans.

\subsection{Boost visual knowledge with big models}

Having highlighted the great importance of visual knowledge in enhancing big AI models, we will next explore the pivotal role that big models play in advancing visual knowledge.

First, big AI models will be an essential cornerstone of visual knowledge. Big models exhibit the unparalleled ability to discern meaningful patterns from large-scale data. Therefore, it is a nature choice to utilize the large-scale learning ability of big models to learn robust visual concepts, and model basic visual relations (such as temporal relations and geometric relations) and operations (such as composition, deformation, and motion).

Second, big AI models will serve as a knowledge source for visual knowledge. Trained on large volumes of text including scientific articles, Wikipedia, books, and other sources of information, large language models have been observed to learn not only contextualized text representations but also significant body of world knowledge and commonsense knowledge~\citep{safavi2021relational}. This suggests the great potential of big models as a knowledge base that could significantly enrich visual knowledge. For example, large language models can help to better capture semantic relationships between concepts, which are often not immediately apparent in visual data; the categorial relationship between ``cat'' and ``animal'' is more readily understood through text. However, unlocking this treasure trove of knowledge is a highly challenging task. The knowledge acquired by big models is deeply hidden within the network parameters, not directly accessible for analysis and reuse. Moreover, the knowledge in big models is not purely factual or reliable; they are intertwined with errors, bias, noise, and trivial patterns. Therefore, advanced techniques~\citep{alkhamissi2022review} for knowledge analysis (identify and localize what knowledge has been acquired by language models), knowledge extraction (extract and represent the knowledge encoded in large models), and knowledge enhancement (validate and refine the extracted knowledge) should be utilized. 

%Their most striking feature is that they model the world, as described by the data on which they were trained. And so really LMs serve as a textual gateway to the universe of knowledge [11, 12], and perhaps should instead be called “language and knowledge” models.

%Thus it is highly desirable to enrich visual knowledge withinternal 

%The journey towards fully leveraging the knowledge within LLMs is complex and ongoing, but it is a crucial step forward in the quest to harness the power of AI for the betterment of society.

Third, big AI models will provide complementary knowledge for visual knowledge. Large language models model the world, as described by the text data. The knowledge acquired from the text data not only enrich, but also complement visual knowledge. For example, some knowledge is hard to learn from visual data, such as human internal thoughts, motivations, and emotions, and commonsense knowledge like ``Beijing is the caption of China''. Similarly,  while a photograph of the Great Wall of China conveys its majesty and scale, it is through textual data that we learn about its historical significance, the reasons for its construction, and its role in Chinese culture. Cognitive studies also suggest visual memory and linguistic memory are not independent but interact with each other in complex ways. Visual memory can be influenced by linguistic information that provides meaning and context to visual stimuli. Linguistic memory can be influenced by visual information that provides imagery and detail to verbal stimuli. As a result, complementing visual knowledge with the knowledge in big models will lead to a more holistic understanding of the world.

%Another study by Loftus et al. showed linguistic influences on visual memory. They asked participants to view a series of pictures of common objects with labels that were either correct or incorrect. They found that the labels affected the accuracy and order of recall of the pictures. Participants were more likely to reproduce the pictures with correct labels than those with incorrect labels. They were also more likely to reproduce the pictures in an order that reflected the conceptual relationships among the labels.

%These studies suggest that visual memory and linguistic memory are not independent but interact with each other in complex ways. Visual memory can be influenced by linguistic information that provides meaning and context to visual stimuli. Linguistic memory can be influenced by visual information that provides imagery and detail to verbal stimuli.

\section{Conclusion}\label{sec:con}
The last decade has witnessed breakthroughs in the field of AI, especially connectionist approaches to long standing challenges around computer vision, natural language processing, speech recognition, and autonomous systems. As the unique product of an era with a treasure trove of data from the internet and increasingly powerful computing resources, big AI models, which assemble the characteristics of both connectionism and scaling law, are swiftly embedding themselves into the fabric of human society and becoming indispensable for scientific discovery. While these developments are nothing short of revolutionary, and have fundamentally altered our way of life, there is a general agreement that this is just the beginning of an AI revolution. However, big AI models still exhibit deficiencies in, for example, transparency, accountability, and symbolic reasoning. Given the significant advantages in the comprehensive modeling of visual concepts, relations, operations, and reasoning, visual knowledge shows the promise of mitigating the shortcomings of existing AI techniques, unlocking the door of the next-generation AI. Starting from reviewing cognitive studies in visual memory and perception, this article introduces the origin and core concepts of visual knowledge. Subsequently, this article retrospects the recent research efforts that are relevant to visual knowledge, along the dimensions of visual concept, visual relation, visual operation, and visual reasoning. Based on the analysis of the current research situation of visual knowledge, we point out the opportunities and challenges it faces in the era of big models, shielding highlight the way of the next-generation AI.

%Over the course of the last decade, AI researchers have made ground-breaking progress in hard and long-standing problems related to machine learning, computer vision, natural language processing, speech recognition,  and autonomous systems. In combination with the availability of planetary scale data from the internet, and increasingly powerful computing resources, AI techniques, especially large AI models, are quickly becoming an integral part of human society as well as an indispensable tool of scientific discovery. As impressive as these developments are, and as much as these technologies have already changed our lives, there is a general agreement that what we see today is just the beginning of an AI revolution. 

% the opportunities and challenges of visual knowledge in the coming big model era, in order to provide reference and guidance for the development of the next-generation of AI. Look ahead, this study demonstrates that visual knowledge will be a key for the next-generation AI.  
% There is also a strong consensus that AI will bring changes that will be much more profound than most other technological revolutions in human history. Depending on the course of this technological revolution will take, AI can either empower individuals and society, creating unimaginable opportunities to improve overall human experience and quality of life, or create the tools to destroy society, enslave individuals, and concentrate power and wealth in the hands of a few.

%This could open the door to new ways of working together and solving problems.

\bibliography{sn-bibliography}

\end{document}